\documentclass[review,12pt]{elsarticle}

\journal{Robotics and Autonomous Systems}
\biboptions{numbers,sort&compress}

\usepackage[utf8]{inputenc}
\usepackage[T1]{fontenc}
\usepackage{amsmath,amssymb,amsfonts}
\usepackage{textcomp}
\usepackage{scalerel}
\usepackage{siunitx}
\usepackage{comment}
\usepackage{pgffor}
\usepackage{pifont}
\usepackage{graphicx}
\usepackage{subcaption}
\usepackage{float}
\usepackage{multirow}
\usepackage{bm}
\usepackage{url}
\usepackage{diagbox}
\usepackage{soul}
\usepackage{wrapfig}
\usepackage[inline]{enumitem}
\usepackage{wasysym}
\usepackage[normalem]{ulem}
\usepackage[dvipsnames,xcdraw]{xcolor}
\usepackage{tikz}
\usepackage[para]{footmisc}
\usepackage[font=small,labelfont=bf]{caption}
\usepackage{algorithm}
\usepackage{algpseudocode}
\usepackage{booktabs}
\usepackage{array}
\usepackage{tabularx}
\usepackage{makecell}
\usepackage[section]{placeins}
\usepackage[hidelinks]{hyperref}
\usepackage[capitalise,nameinlink]{cleveref}
\usepackage{graphicx}

\usetikzlibrary{positioning,fit,calc,backgrounds,shadows,arrows.meta}

\definecolor{customgreen}{HTML}{c0e9d2}

\newcommand{\cmark}{\ding{51}}
\newcommand{\xmark}{\ding{55}}
\newcolumntype{Y}{>{\centering\arraybackslash}X}

\usepackage{xcolor}
\usepackage[normalem]{ulem} 

\begin{document}

\begin{frontmatter}


\title{Fly, Track, Land: Infrastructure-less Magnetic Localization for Heterogeneous UAV--UGV Teaming}

\author[unipg,ctu]{Valerio Brunacci}
\author[eth]{Davide Plozza}
\author[unipg]{Alessio De Angelis}
\author[eth]{Michele Magno}
\author[eth]{Tommaso Polonelli}

\affiliation[unipg]{organization={Department of Engineering, University of Perugia},
            city={Perugia},
            postcode={06125},
            country={Italy}}
\affiliation[ctu]{organization={Multi-Robot Systems Group, Faculty of Electrical Engineering, Czech Technical University in Prague},
            city={Prague},
            postcode={16627},
            country={Czech Republic}}
\affiliation[eth]{organization={Department of Information Technology and Electrical Engineering (D-ITET), ETH Zurich},
            city={Zurich},
            postcode={8092},
            country={Switzerland}}

\begin{abstract}
{\color{orange}\color{black}
Persistent air ground robot teams require nano UAVs that can leave a mobile UGV, perform short scouting or inspection tasks, and reliably return to a compact landing interface. This final docking phase remains difficult: it demands centimeter-scale relative localization on a moving platform, while nano UAVs provide only a few grams of payload and severe onboard sensing, power, and computation constraints.
We address this problem with an infrastructure-less magneto inductive anchor-tag localization system for heterogeneous UAV--UGV teaming.
Unlike passive magnetic docking aids or approaches exploiting environmental magnetic features, the proposed system actively generates a frequency multiplexed AC magnetic field on the UGV and uses it as an onboard localization reference for closed loop  flight.
The nano UAV carries only a lightweight passive receive coil and estimates its 3D position directly in the UGV frame, providing a local close range reference for hovering, tracking, and landing without external anchors, visual fiducials, GNSS, or motion capture online. 
The magnetic estimate is fused with the native onboard sensing stack, enabling operation under nano UAV SWaP constraints. 
Experiments with a Unitree A1 quadruped and a Crazyflie nano UAV, evaluated against motion capture ground truth, show centimeter level accuracy in static hovering and landing and approximately 8--11 cm RMSE during UGV motion, while the flow only baseline frequently violates the safety bound and fails the task. 
}
\end{abstract}



\end{frontmatter}

\section{Introduction}\label{sec:introduction}
With the increasing popularity of multi-robot swarms, collaborative exploration and mission planning have gained significant scientific attention \cite{zhou2023racer, lanvca2025probabilistic}. However, current research predominantly focuses on homogeneous robot teams \cite{xu2022omni, javed2024state}, leaving substantial opportunities for advancing collaboration between ground and aerial platforms \cite{cai2023energy}. Recently, heterogeneous robotic teams have attracted growing interest for applications where complementary capabilities can be effectively leveraged~\cite{tranzatto2022cerberus, zhang2024heterogeneous}. In particular, unmanned aerial vehicles (UAVs) \cite{niculescu2025ultra} have emerged as agile and easily deployable aerial perception agents that can complement unmanned ground vehicle (UGVs) such as quadrupeds or wheeled robots \cite{yang2021survey}. For instance, nano-scale palm-sized UAVs can access elevated vantage points or confined spaces beyond the reach of ground robots, while larger UGVs can transport heavier payloads and host more powerful sensing and computing resources for mapping and navigation~\cite{ribeiro2021unmanned}. This synergy combines the mobility and rapid scouting capability of aerial robots with the endurance and sensing capacity of ground platforms~\cite{tranzatto2022cerberus}. Leveraging the complementary characteristics of heterogeneous robotic swarms can substantially enhance exploration performance and mission efficiency. In such systems, UGVs provide payload capacity and long operational endurance, while aerial robots contribute agility and maneuverability. For example, a nano UAV deployed from a legged robot can scout difficult-to-access areas \cite{pourjabar2023land} and subsequently return to the UGV for data transfer or battery recharging, enabling more informed and globally consistent task planning.

Such marsupial robot configurations have been proposed to maximize mission coverage and efficiency~\cite{tranzatto2022cerberus}. Indeed, recent efforts integrated a micro-drone with a quadruped as a mobile launch/landing pad~\cite{tranzatto2022cerberus}, and NASA’s \emph{Ingenuity} Mars helicopter (a \qty{1.8}{\kilogram} UAV) demonstrated the benefits of aerial ground cooperation by scouting ahead of the Perseverance rover~\cite{chien2024exploring}. These examples motivate our focus on enabling a nano UAV to precisely land on a mobile UGV – a capability that can significantly enhance the deployability of future robot teams in the field \cite{zhou2022swarm, niculescu2025ultra}.

\begin{figure}[H]
    \centering
    \includegraphics[width=\linewidth,height=0.62\textheight,keepaspectratio]{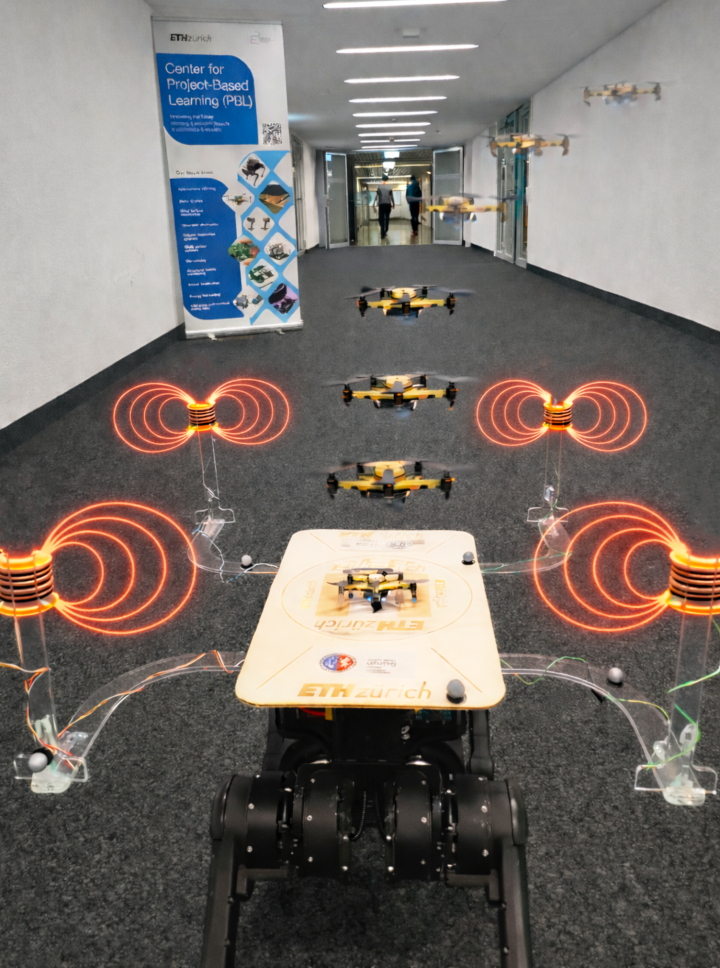}
    \caption{Representative picture of a nano-size UAV autonomously landing on a moving legged robot and GNSS-denied environment. The infrastructure-free localization systems relies on onboard sensors, combining IMUs, optical-flow camera, UWB, and the MI system. The four reference coils are represented by the red lines with a magnetic dipole shape. The landing deck can be used to recharge the UAV's batteries.}
    \label{fig:catchy_figure}
\end{figure}

A fundamental challenge to enable heterogeneous robotic collaborations relies in achieving robust relative localization between the nano UAV and the moving UGV with the accuracy needed for docking \cite{niculescu2025ultra}. 
Conventional solutions such as UWB ranging and GNSS (including RTK variants) are widely used to provide global or relative positioning over meter-scale workspaces~\cite{schindler2023relative, polonelli2022performance, mikhaylov2023toward}. 
Moreover, GNSS can be unavailable or unreliable indoors, in urban canyons, underground, or in planetary-like scenarios, motivating infrastructure-less relative sensing for close-proximity interaction~\cite{Liu2025}. 
In practice, their achievable accuracy strongly depends on deployment and operating conditions (e.g., anchor geometry and calibration for UWB, or correction availability for GNSS), and their role is often best suited to coarse acquisition, navigation, and rendezvous at longer ranges.
However, during the \emph{final docking phase} onto a compact, mobile interface (tens of centimeters), even residual decimeter-level relative errors can lead to missed contact or unsafe interactions, motivating sensing modalities that are specifically effective at close range~\cite{pourjabar2023land}.

Vision-based relative localization can provide high precision by detecting fiducial markers or recognizable features on the UGV~\cite{lin2021low, niu2021vision, ladosz2024autonomous}. 
However, vision performance depends on favorable illumination and unobstructed line-of-sight, and can degrade in darkness, dust, smoke, or cluttered scenes -- conditions frequently encountered in realistic deployments~\cite{tranzatto2022cerberus}. 
Moreover, ultra-lightweight nano UAVs face strict SWaP limits, which restrict camera choice, onboard compute, and the complexity of perception pipelines~\cite{niculescu2025ultra}. 
In summary, existing modalities offer strong advantages at different scales, but none alone consistently satisfies the combined requirements of \emph{infrastructure-less}, close range, centimeter-accurate docking onto a moving UGV in challenging environments.

{\color{orange}\color{black}
To address the final stage of UAV--UGV docking, we propose an active magneto inductive (MI) anchor-tag relative localization system. The key idea is to make the UGV a mobile local reference frame: four frequency-multiplexed transmitting coils mounted on the quadruped generate a structured AC magnetic field, while the
nano UAV carries only a lightweight passive receiving coil. The received signal is processed onboard the UAV to estimate its 3D position directly in the UGV frame, providing a close range absolute relative measurement that does not depend on external infrastructure, visual markers, GNSS, motion capture, or pre-existing
environmental magnetic maps.

We adopt a hierarchical view of relative localization. Mid and long range acquisition can be provided by existing modalities such as GNSS when available, RF/UWB ranging, visual-inertial localization, or LiDAR/SLAM-based navigation. These sensing layers are valuable for coarse rendezvous and mission-level navigation, but the final centimeter level docking phase requires a local relative reference whose accuracy is comparable to the size of the nano UAV and the landing interface. The proposed MI layer is therefore not intended to
replace mid-range localization systems. Instead, it is specialized for the last-meter phase, where the UAV must hover, track, and land on a compact moving base under severe size, weight, and power constraints.

Unlike passive magnetic docking aids or approaches that exploit ambient magnetic field anomalies, the proposed system actively generates a controllable AC magnetic reference on the UGV and solves the corresponding inverse localization problem onboard the UAV. To the best of our knowledge, this is the first demonstration of onboard active magnetic anchor-tag relative localization for closed loop nano UAV hovering, tracking, and landing on a mobile quadruped, independent of environmental magnetic field maps or other external environmental measurements.

Our custom lightweight sensor, namely "MagneticDeck" performs signal acquisition and spectral extraction of the four anchor frequencies, applies per-anchor calibration, and, then, the UAV onboard MCU solves the inverse magnetic positioning problem using a Nelder--Mead optimizer. 
The resulting position estimate is fused into the native onboard flight estimator together with inertial, optical-flow, and altitude measurements, enabling closed loop control without offboard computation.

We validate the system experimentally using a Unitree A1 quadruped as the mobile anchor platform and docking station, and a Crazyflie 2.1 nano UAV as the airborne tag. Motion capture is used only as ground truth for evaluation, not as an online localization source. 
In static hovering and landing, the system achieves centimeter level accuracy and reliable touchdown on the docking pad. 
In dynamic scenarios, where the UGV translates while the UAV tracks or lands, the proposed magnetic aided estimator maintains errors in the order of 8--11 cm, whereas the native flow-only baseline frequently violates the safety bound and fails the task.

In summary, this paper makes the following contributions:
\begin{enumerate}
    \item We introduce a magnetic anchor-tag relative localization architecture for infrastructure-less nano UAV docking on a mobile quadruped. The UGV acts simultaneously as a mobile magnetic reference frame and docking station, while the UAV carries only a lightweight passive receiver.

    \item We demonstrate closed loop hovering, tracking, and landing of a Crazyflie-class nano UAV on a Unitree A1 quadruped using onboard magnetic relative localization fused with the native flight estimator.

    \item We show that active AC magnetic localization can be implemented within nano UAV SWaP constraints by running the full signal extraction,calibration, and inverse position estimation pipeline onboard the
    Crazyflie microcontroller.

    \item We provide real world validation against motion capture ground truth in static and dynamic UGV scenarios, together with a discussion of the current operational envelope and failure modes, including UGV
    attitude changes.
\end{enumerate}
}

\section{Related Work} \label{sec:related}

Combining aerial and ground robots into cooperative teams has gained significant attention for missions that demand both endurance and mobility \cite{zheng2025aage}, such as planetary exploration~\cite{chien2024exploring}, subterranean mapping~\cite{tranzatto2022cerberus}, search and rescue~\cite{ribeiro2021unmanned}, and infrastructure inspection~\cite{Ochoa2022}. Ground robots—particularly legged platforms like quadrupeds—offer payload capacity, energy efficiency, and stable locomotion across uneven terrain \cite{ha2025learning}. However, their sensing capabilities are limited by their low vantage point and occlusions near ground level  \cite{anymal_parkour, agarwal2023legged}. This restricts their ability to perceive upcoming terrain features, particularly negative obstacles such as drop-offs or gaps, or to navigate around obstacles beyond their immediate field of view.
Prior works have therefore explored the use of aerial companions to compensate for these limitations, enabling real time overhead imagery \cite{zheng2025aage}, traversability assessment, or 3D mapping from dynamically changing viewpoints \cite{chien2024exploring,niu2021vision}.
Such a configuration enables truly collaborative autonomy, where the UAV enhances the perception and planning of the UGV, and the UGV in turn provides computational support, energy, or physical transport for the UAV \cite{niu2021vision}. Numerous works have demonstrated UAV–UGV teams in exploration, mapping, and target tracking \cite{Liu2022_cooperative_robotics_review_JIRS}. A common framework is a UGV serving as a mobile base or “hub” for UAVs \cite{tokekar2016_symbiotic_sensor_planning}. For example, in search \& rescue scenarios, a UGV can carry extra batteries or sensors, and a UAV deploys from it to survey areas and then returns to the UGV for data transfer or charging \cite{polonelli2022open}. 
This requires the UAV to localize relative to the UGV \emph{without} relying on external infrastructure.  Space robotics offers a compelling real world precedent \cite{chien2024exploring}: NASA’s Perseverance rover and its aerial companion, the \emph{Ingenuity} helicopter, demonstrated that heterogeneous robot teams can outperform monolithic systems by combining sensing perspectives and operational ranges \cite{chien2024exploring}. Inspired by this paradigm, recent terrestrial efforts have explored tightly coupled UAV--UGV systems for deployment in GNSS-denied and infrastructure-free environments \cite{yue2021tightly,Liu2025}. 

However, these missions face multiple challenges. GNSS and GPS-based localization, while widely used, are often unavailable or unreliable in cluttered urban canyons, indoors, underground, or on other planets \cite{Liu2025}. Even when GNSS is available, its typical accuracy (on the order of \qtyrange{1}{3}{\meter}) is inadequate for high-precision coordination or docking maneuvers \cite{mikhaylov2023toward}. To overcome these challenges, a wide range of approaches have been proposed for relative localization and docking in air ground systems, which we explore in detail in the following subsections and compare with our work in Table \ref{tab:uav_ugv_landing_comparison}. 

\subsection{Relative Localization for Teaming Heterogeneous Robots and Docking}\label{sec:relativelocalization}
A wide body of work explores relative localization to support collaboration \cite{zhang2024heterogeneous, niculescu2025ultra, schindler2023relative}. Vision-based methods are among the most accurate \cite{ferreira2025heterogeneous}: fiducial markers (e.g., AprilTags~\cite{krogius2019flexible}) or model based object detection can yield centimeter-scale pose estimates under ideal conditions. However, their dependence on line-of-sight, good illumination, and high camera frame rates makes them unreliable in cluttered or fast-changing environments \cite{Salagame2022}.
Moreover, the ultra-lightweight UAVs suitable for scouting from a legged robot—such as Crazyflie-class nano-drones—face extreme constraints in computation, payload, and sensor quality, which limit their ability to run conventional perception pipelines \cite{niculescu2025ultra,schindler2023relative}. Onboard cameras with wide-angle lenses or dedicated neural detectors partially address this \cite{ladosz2024autonomous}, but often exceed the SWaP (size, weight, and power) capabilities of nano UAVs \cite{niculescu2025ultra}.

Radio-frequency methods, especially UWB \cite{polonelli2022performance}, offer robustness to lighting and occlusion and have been used in both fixed-anchor and peer-to-peer setups~\cite{niculescu2025ultra,schindler2023relative}. UWB allows fully mobile teams to localize one another without infrastructure \cite{schindler2023relative}, but the achievable accuracy (typically \qtyrange{10}{30}{\centi\metre}) falls short for tasks requiring precise relative pose for landing and docking \cite{yu2025quadrotor}. This limitation is visible even on Crazyflie-scale persistent autonomy: Nguyen \emph{et al.}~\cite{Nguyen2022PersistentCharging} rely on an anchor-based loco positioning system for tracking, but switch to a multi-ranger (laser ToF) deck during the landing phase to reduce horizontal error and achieve reliable pad contact.
Moreover, multiple antennas or cooperative estimation algorithms are often needed to infer full 6-DOF pose \cite{schindler2023relative}, adding system complexity and calibration burden.

{\color{orange}\color{black}
This limitation is particularly relevant for nano UAVs, whose characteristic size is often around or below 10\,cm and whose payload, sensing, and power budgets are severely constrained \cite{bonato2023monocular,pourjabar2023land,zango2026nano}. In confined UAV settings, UWB based localization methods typically report decimeter-level accuracy rather than consistent sub 10\,cm performance:
average localization errors below 0.2\,m have been reported in extremely confined environments, while controlled indoor evaluations reported average errors of 6.95\,cm in a $3\times3$\,m area and 9.34\,cm in a
$3\times12$\,m area, highlighting the sensitivity of realized accuracy to deployment geometry and environment
\cite{yang2022confined,sorescu2024uwb}. Representative nano UAV and swarm systems based on UWB report localization RMSEs of 15.3--27.8\,cm in field deployments, while UWB based 3D node localization from a nano UAV remains bounded within 28\,cm even after dedicated ranging-error correction \cite{pourjabar2023land,niculescu2023iotj}. More broadly, UWB-only or UWB-dominant systems often require explicit NLOS mitigation, learned range-error correction, or tightly coupled filtering to approach
decimeter-level robustness \cite{yang2023svm,coene2024larc,fontaine2024transfer,jia2024cdf}.

Accordingly, we consider UWB and related RF systems as useful mid-range acquisition layers rather than as final-contact localization solutions for the specific nano UAV docking problem addressed here.
For platforms whose dimensions are comparable to the required error bound, close range UWB is operationally useful mainly when combined with bias calibration, sensor fusion, constrained geometry, or complementary sensing. This motivates the proposed active magnetic layer, which is specialized for the final close-proximity phase rather than for global or long-range navigation.

}

Some systems employ an “on-body anchor” strategy, where the UGV aids localization by carrying markers and beacons \cite{Liu2025}. In~\cite{de2024uav}, for example, a UGV emits visual or UWB signals for UAV tracking and homing. Cooperative localization filters have also been proposed, combining the UGV’s odometry with the UAV’s observations to maintain a shared relative estimate~\cite{Liu2025}. These approaches remove the need for fixed infrastructure, but still face accuracy and robustness trade offs in dynamic or cluttered scenes.

Current trends point toward sensor fusion: combining modalities like vision, RF ranging, and inertial data to maximize robustness across environments~\cite{airgroundreview2020, Liu2025}. Recent reviews~\cite{airgroundreview2020} converge on a key insight: no single sensing modality suffices for robust air ground coordination across all conditions. Instead, fusing complementary sensors—each optimized for different stages or scales of the task—is the emerging trend \cite{nguyen2021viral}. However, integrating such systems on small-scale UAVs remains a challenge due to tight SWaP constraints \cite{niculescu2025ultra}. 

As robotic collaboration expands toward larger, autonomous teams—such as drone–quadruped formations in underground or extraterrestrial missions—there is growing demand for centimeter-precision, infrastructure-free localization methods that are resilient, low-power, and compatible with ultra-light aerial platforms \cite{niculescu2025ultra}. As visible from \Cref{tab:uav_ugv_landing_comparison}, this paper is bridging this gap enabling close aerial-ground cooperation by introducing MI localization as a complementary sensing modality for close-proximity UAV–UGV teaming, particularly suited to the final stages of coordinated tasks such as landing, battery charging, mapping, tracking, or deployment. Compared with previous methods listed in \Cref{tab:uav_ugv_landing_comparison}, our work sidesteps the limitations of common localization approaches, e.g., GNSS, camera-based, and UWB range resolution, while being lightweight enough for integration into centimeter-scale UAVs. It supports a new class of tightly integrated, heterogeneous robotic teams suitable for autonomous deployment in unstructured and GNSS-denied environments.

A particularly challenging case of UAV–UGV cooperation is autonomous landing of a UAV on a \emph{moving} UGV.
This is essentially a precision pursuit-evasion problem: the UAV must not only track the UGV’s pose but also descend and land, often within a limited “landing zone” area on the UGV. A critical requirement here is robust disturbance rejection and trajectory planning, where the UAV needs to anticipate the UGV’s motion.
Many solutions maintain a constant offset or approach from behind to account for UGV forward motion \cite{MovingPlatformApproach2017}. Approaches like Baca \cite{Baca2019} used a fisheye camera for a wide view of the moving target and a Kalman filter to estimate its trajectory, then guided the UAV in for landing.
Mobile docking requires careful software solutions: the literature emphasizes robustness for uncertainty aware path planners (ability to land despite disturbances or sudden UGV maneuvers) as a primary metric for success \cite{deng2023self}.
The state of the art now includes demonstrations of UAVs landing on UGVs that are not only moving but possibly adapting – e.g., a UGV with a self-leveling platform to assist the UAV \cite{alghanim2020comparison}. Still, there are open research gaps: multi-UAV, single-UGV coordination (scheduling which drone lands when, to avoid conflicts) and performing this reliably without GNSS or motion capture \cite{Liu2025,lan2020induction}. A body of work focuses on persistent docking of UAVs for battery recharge or payload transfer \cite{lan2020induction}. These systems often employ custom fiducials for final alignment \cite{serra2016_landing_moving_target_ibvs}.
For instance, a number of recent designs use wireless charging pads that guide the UAV visual markers into a receptacle for inductive charging \cite{yuan2022high}. The systems in \cite{yuan2022high, nguyen2023novel} combines downward camera-based marker detection with precision flight control to land on a charging pad, achieving reliable contact for recharging \cite{lan2020induction}. The emphasis in such works is on ensuring physical alignment (to connect power terminals or coils) with a sub centimeter level precision, needed for accurate landing and inductive battery recharging on a moving target \cite{lan2020induction}. IR-LOCK \cite{khoirunnisa2023implementation} is one commercial example, using an IR beacon on the pad and a filtered IR camera on the drone to zero in on the hotspot – effective indoors, but essentially a homing sensor rather than a general localization solution. Mechanical docking stations with funnels or self-aligning mechanisms have also been developed, reducing the required landing precision at the cost of added weight on the docking platform \cite{sun2025parallel}. These solutions highlight practical infrastructure for autonomy, but they typically assume a mostly static or pre-known landing site.
Unlike our scenario of mobile landing on a UGV, standard docking pads do not estimate the relative 3D pose during approach – they often presuppose the UAV navigates itself near the pad using other means, then use short range guidance (vision or IR) for the final meter. 
\textcolor{black}{Some docking systems also exploit permanent magnets or electromagnets to provide short range attraction and self-alignment (e.g., \cite{yu2025quadrotor}); however, these approaches function as passive alignment aids rather than structured sensing systems, providing short range attraction but no measurement model for 3D pose estimation, a fundamental distinction from the active AC MI system proposed in this work.}
In summary, vision-based landing offers high precision but can be fragile to lighting  and line of sight \cite{Salagame2022} and require heavy processing; range and RF methods are more tolerant to environment but provide less precision without infrastructure.
None of these categories simultaneously satisfy the centimeter-scale accuracy, infrastructure independence, and strict payload/power envelope demanded by nano UAV platforms teamed with moving UGVs, motivating the magneto inductive localization alternative explored in this work.

\subsection{magneto inductive and near field Magnetic Localization}\label{sec:magneto inductive}

MI localization leverages low-frequency magnetic fields that decay with $1/r^3$, making them inherently robust to multipath effects and largely unaffected by non-ferromagnetic occlusions~\cite{Abrudan2015,Markham2012}. Most systems model the transmitters as magnetic dipoles—small coils generating predictable axial and transverse field patterns in free space \cite{Chavda2023}. This dipole approximation is accurate when the receiver is located several coil diameters away but becomes less reliable at close ranges. For applications such as drone docking at $\sim$\qty{0.5}{\meter} with coil diameters of \qtyrange{5}{10}{\centi\meter}, the system operates near the limits of dipole model validity~\cite{DeAngelis2017}. Mutual inductance models or geometric corrections may be necessary to achieve sub-centimeter accuracy in fields spanning $\sim$\qty{30}{\centi\meter}.

3D localization requires distinguishing fields from multiple transmitters. We adopt frequency-division multiplexing (FDM), in which each coil broadcasts a sinusoidal signal at a distinct frequency. The UAV onboard magnetometer (see \Cref{fig:hardware_composite}) measures the superimposed field, and spectral decomposition (e.g., FFT) isolates individual coil contributions~\cite{Chavda2023}. FDM enables continuous sampling but requires careful frequency spacing to avoid mixing artifacts~\cite{FDMcrosstalk2019}. Coil LC resonance is tuned to maximize field strength (high Q factor), and signal amplitudes are calibrated to avoid saturating the receiver’s ADC, especially at close proximity. Position estimation relies on solving an inverse magnetic problem: given known coil positions and field vectors at the receiver, compute the 3D receiver position (and potentially orientation). Simpler systems use only range estimation via the $1/r^3$ decay, but this ignores orientation and cross-axis coupling. More general solutions involve nonlinear optimization, such as gradient descent (e.g., Nelder–Mead), Gauss-Newton methods, or global search techniques~\cite{MedMagTracker2016}. Filters like the Unscented Kalman Filter (UKF) have been applied to track coil pose in real time, enabling sub-centimeter accuracy~\cite{DeAngelis2017}. Our implementation simplifies the problem by using IMU-based attitude estimation from the Crazyflie, reducing magnetic inversion to position only estimation.

Notably, MI localization is resilient to lighting conditions and visual occlusion—smoke, walls, or darkness have negligible impact on low-frequency magnetic fields. Unlike RF, MI systems are immune to multipath and reflections from materials like wood or concrete, which have relative permeability close to unity~\cite{Chavda2023}. However, nearby ferromagnetic materials or conductors can introduce bias or attenuation due to eddy currents. We mitigate this with in-situ calibration and analog/digital filtering focused on narrow frequency bands of interest. Motor EMI from the drone is filtered out due to its distinct frequency content~\cite{MEGCoilCalib2019}. Our system complies with EMC safety standards, as our frequencies (tens of kHz) and field strengths are comparable to wireless charging systems and confined within the near field region.

MI tracking has proven effective in specialized applications such as biomedical implant localization and AR/VR headsets, achieving millimeter-level accuracy in $\sim$\qtyrange{30}{50}{\centi\meter} volumes~\cite{DeAngelis2017}. More recently, MI systems have been deployed in mining and underwater robotics where RF fails~\cite{Chavda2023}. In mobile robotics, however, their use for aerial docking or robot teaming remains limited. Our system demonstrates MI localization as a viable solution for nano UAV landing in indoor or GNSS-denied settings, offering a precise, infrastructure-free alternative to visual or UWB based approaches under severe SWaP constraints.

\begin{table*}[t]
  \centering
  \caption{Comparison of UAV-UGV landing and cooperative systems in recent literature.}
  \label{tab:uav_ugv_landing_comparison}
  \resizebox{\linewidth}{!}{\begin{tabular}{@{}p{1.45cm}p{3.55cm}cc>{\centering\arraybackslash}p{1.9cm}>{\centering\arraybackslash}p{2.0cm}>{\centering\arraybackslash}p{2.1cm}>{\centering\arraybackslash}p{1.1cm}@{}}
      \toprule
      \textbf{Paper} &
      \textbf{Localization Method} &
      \textbf{Infrastructure-less} &
      \textbf{Moving Landing} &
      \textbf{Landing metric} &
      \textbf{UAV-UGV Teaming} &
      \textbf{Field Test} &
      \textbf{nano UAV} \\
      \midrule

      \cite{Liu2025} &
      UWB+IMU+Vision &
      \xmark &
      \cmark &
      - &
      \xmark &
      Outdoor &
      \xmark \\

      \cite{de2024uav} &
      UWB ranging (2 UAVs) &
      \cmark &
      \xmark & \qty{14}{\cm} &
      \cmark &
      Outdoor &
      \xmark \\

      \cite{polonelli2022open} &
      GNSS + UWB &
      \xmark &
      \xmark &
      $\pm$\qty{10}{\cm} &
      \xmark~(static sensors) &
      Outdoor &
      \xmark \\

      \cite{nguyen2024landing} &
      \makecell[l]{Vision + motion} &
      \cmark &
      \cmark &
      $\pm$\ang{5} &
      \xmark &
      Simulation &
      \xmark \\

      \cite{ladosz2024autonomous} &
      Vision &
      \cmark &
      \cmark &
      80\% SC &
      \xmark &
      \mbox{Sim. + Indoor} &
      \cmark \\

      \cite{zheng2025aage} &
      SLAM (shared map.) + manual &
      \xmark &
      \xmark &
      - &
      \cmark &
      Outdoor &
      \xmark \\

      \cite{niculescu2025ultra} &
      UWB + Vision &
      \xmark &
      - &
      \qty{30}{\cm} &
      \xmark &
      Indoor &
      \cmark \\

      \cite{schindler2023relative} &
      UWB &
      \cmark &
      - &
      \qty{39}{\cm} &
      \xmark &
      Indoor &
      \cmark \\

      \cite{zhang2024heterogeneous} &
      restricted &
      \cmark &
      \xmark &
      - &
      \cmark~(rover) &
      Moon/Mars &
      \cmark \\

      \cite{niu2021vision} &
      GNSS + Vision &
      \xmark &
      \cmark &
      \mbox{\qtyrange{20}{60}{\cm}} &
      \cmark &
      Simulation &
      \xmark \\

      \cite{brunacci2023fusion} &
      UWB + MI &
      \xmark &
      \xmark &
      \qty{7}{\cm} &
      \xmark &
      Outdoor &
      \xmark \\

      \cite{yu2025quadrotor} &
      MI + Vision &
      \cmark &
      \xmark &
      $\sim$\qty{1}{\cm} &
      \xmark &
      \mbox{Sim. + Indoor} &
      \xmark \\

      \cite{yue2021tightly} &
      Vision &
      \xmark &
      \xmark &
      - &
      \cmark &
      Indoor &
      \xmark \\

      \cite{lan2020induction} &
      MI &
      \cmark &
      \xmark &
      - &
      \xmark &
      - &
      \xmark \\

      \cite{deng2023self} &
      Mechanical &
      \xmark &
      \cmark &
      \qty{4.5}{\cm} &
      \cmark &
      Indoor &
      \xmark \\

      \cite{ferreira2025heterogeneous} &
      Vision &
      \xmark &
      \cmark &
      $\pm$\qty{20}{\cm} &
      \cmark~(boat) &
      Outdoor &
      \cmark \\

      \midrule

      \textbf{This work} &
      \textbf{MI} &
      \textbf{\cmark} &
      \textbf{\cmark} &
      \textbf{7--11~cm} &
      \textbf{\cmark} &
      \textbf{Indoor} &
      \textbf{\cmark} \\

      \bottomrule
    \end{tabular}}
\end{table*}

\section{Magnetic Model \& Estimator}
\label{sec:mag_model}

We consider a heterogeneous air ground system composed of a quadruped UGV equipped with $N=4$ transmitting coils (anchors) and a nano UAV carrying a single receiving coil (tag). Let $\{\mathcal{B}\}$ denote the inertial frame attached to the UGV (anchor frame), and $\{\mathcal{T}\}$ the body frame rigidly attached to the UAV. The system geometry is defined by the fixed configuration of the anchors on the ground unit. The $i$-th transmitting coil is located at a known position $\mathbf{p}_i^{\mathcal{B}}$ relative to the UGV center. To simplify the setup, the anchors are mounted with their magnetic axes strictly perpendicular to the UGV horizontal plane; consequently, the orientation vector $\mathbf{u}_i$ of each transmitter is fixed and aligned with the vertical axis of the reference frame (i.e., $\mathbf{u}_i \parallel \mathbf{z}_{\mathcal{B}}$). On the receiving side, the tag coil is rigidly mounted to the UAV airframe. Its orientation is defined by the local unit normal $\mathbf{n}^{\mathcal{T}}$, which is constant in the body frame $\{\mathcal{T}\}$. However, since the coil rotates integrally with the drone, its effective orientation in the anchor frame, denoted as $\mathbf{n}^{\mathcal{B}}$, varies over time. This vector is continuously reconstructed using the UAV's onboard attitude estimate via the rotation matrix $\mathbf{R}_{\mathcal{T}}^{\mathcal{B}}$ provided by the flight controller (i.e., $\mathbf{n}^{\mathcal{B}} = \mathbf{R}_{\mathcal{T}}^{\mathcal{B}} \,\mathbf{n}^{\mathcal{T}}$).
{\color{orange}\color{black}
The formulation above compensates the attitude of the UAV-side receiving coil through the onboard attitude estimate. Therefore, roll and pitch motions of the nano UAV do not invalidate the measurement model as long as the flight controller provides an accurate estimate of the receiver-coil normal. Conversely, the current implementation assumes that the UGV anchor frame is fixed with respect to the estimator frame. In particular, the anchor positions $\mathbf{p}_i^B$ and dipole axes $\mathbf{u}_i^B$ used by the inverse solver are initialized in the nominal UGV body frame $B_0$ and are not updated online according to the UGV attitude. If the true UGV body frame is rotated by $R_{B_t}^{B_0}$ with respect to the nominal frame, the physically generated field is produced by the transformed anchors
\[
\mathbf{p}_{i,t}^{B_0}=R_{B_t}^{B_0}\mathbf{p}_i^B,
\qquad
\mathbf{u}_{i,t}^{B_0}=R_{B_t}^{B_0}\mathbf{u}_i^B,
\]
whereas the estimator still evaluates the model using $(\mathbf{p}_i^B,\mathbf{u}_i^B)$. This mismatch is interpreted by the position only solver as a translational displacement and can therefore introduce systematic bias during uncompensated UGV yaw, pitch, or roll. This limitation is quantified in Sec.~\ref{sec:results_limitations}.
}

{\color{orange}\color{black}
The estimation problem consists of determining the unknown position of the tag 
$\mathbf{x}^{\mathcal{B}} \in \mathbb{R}^3$ expressed in the anchor frame 
$\{\mathcal{B}\}$, relying on the magnetic interaction between the ground based 
sources and the receiver. Selecting an appropriate magnetic field model 
is a design choice that balances physical fidelity, computational cost, and the 
operating envelope of the application. In magnetic positioning systems 
(MPS), where small coils generate and sense fields to infer range, bearing, or 
pose at short to mid distances, the sources are typically compact 
\cite{DeAngelis2017}, the environment is weakly dispersive, and the excitation 
frequencies are selected to remain in the near field regime while maintaining 
sufficient signal to noise ratio (SNR). Under these conditions, different modeling 
layers can be invoked with increasing levels of fidelity and computational 
complexity \cite{DeAngelis2017}.

The magnetic forward model used by the estimator must balance three requirements: 
physical consistency in the operating regime, sufficient spatial informativeness 
for inverse localization, and real time evaluability on the nano UAV 
microcontroller. In principle, the field generated by the transmitting coils could 
be described with increasing levels of fidelity. A full Maxwell/FEM formulation 
would capture radiation effects, induced currents, complex geometries, and 
material properties of the surrounding structure. However, this level of 
description requires detailed boundary conditions and material parameters of the 
UGV body and is not compatible with repeated evaluations inside an onboard 
nonlinear optimizer. For this reason, full-wave or FEM-based electromagnetic 
models are more appropriate for offline characterization and platform specific 
distortion analysis than for the runtime estimator used in this work 
\cite{van_Oosterhout2022_FEM}.

A magnetoquasistatic Biot--Savart model of the finite coils would provide a more 
accurate description of the local field generated by the actual windings. This 
model naturally accounts for the finite coil geometry and is well suited to 
compact conductors operating in the near field. However, it still requires 
numerical integration or discretization of the coil geometry at each candidate 
receiver position, making it more expensive than closed form compact source 
models. Moreover, comparative studies on 3D magnetic localization models show 
that, in calibrated and bounded positioning volumes, the practical localization 
gain obtained by replacing a dipole approximation with more detailed field or 
coupling models is not always large enough to justify the additional runtime cost 
in embedded closed loop applications 
\cite{DeAngelis2017,De_Angelis2017_comparison_dipole_mutual_B_models}.

We therefore adopt the magnetic dipole model as a compact source 
approximation. 
This choice is not intended to provide a complete electromagnetic 
simulation of the whole UAV--UGV system. Instead, it provides the simplest forward 
model that preserves the dominant spatial structure required for localization: 
the inverse-cubic decay with distance and the angular dependence between the 
transmitter dipole, the Tx--Rx direction, and the receiving-coil normal. The 
operating frequency is in the low frequency near field regime, where the 
electromagnetic wavelength is orders of magnitude larger than the robot scale 
workspace, so radiative effects are negligible for the purpose of relative 
positioning. Moreover, the transmitting coils are small with respect to the 
docking workspace, and the receiver operates at distances of several coil 
dimensions during hovering, tracking, and landing. Under these conditions, the 
dipole model provides a physically consistent and computationally lightweight 
approximation of the field shape used by the inverse solver 
\cite{De_Angelis2017_comparison_dipole_mutual_B_models}.

Importantly, the estimator does not rely on an absolute level prediction 
of the received voltage. 
The dipole model is used after an in-situ per-anchor 
calibration step that maps the raw FFT peak amplitudes to the corresponding model 
amplitudes at a known reference pose. The calibration coefficients absorb static 
scale factors such as coil manufacturing tolerances, resonance gain, analog 
amplification, ADC scaling, and constant per-channel coupling differences. 
Therefore, the online inverse problem mainly exploits the spatial variation of 
the calibrated amplitudes across the four anchors, rather than requiring an exact 
absolute prediction of the magnetic flux density everywhere in the workspace.

This modeling choice is also aligned with the constraints of the robotic task. 
The proposed system targets short range relative docking inside a bounded 
workspace, not general purpose magnetic field mapping in arbitrary environments. 
The dipole model admits a closed form evaluation, enables multiple candidates 
evaluations within each Nelder--Mead update, and allows the full signal 
extraction, calibration, and inverse position estimation pipeline to run onboard 
the Crazyflie MCU at the control relevant update rate. 
More complex Full 
Maxwell/FEM or Biot--Savart models could improve field fidelity offline, but 
would increase the state estimation latency and implementation complexity without 
changing the main closed loop requirement of the present system: providing a 
stable, calibrated, and sufficiently accurate relative position measurement for 
final UAV--UGV docking.

Nearby conductive or ferromagnetic structures, such as the UGV body or payload, 
may perturb the ideal dipole field through attenuation, eddy current effects, or 
local changes in magnetic permeability. In our setup, these effects are mitigated 
by two practical factors: first, the dominant metallic structures are rigidly fixed 
with respect to the anchor frame and therefore introduce mostly repeatable, 
platform-specific biases; second, the generated magnetic field amplitudes are small, 
making induced perturbations difficult to observe and measure at the system level. 
As a result, no measurable degradation of the final hovering, tracking, or landing 
performance was observed in the tested configuration, and the residual static bias 
is partially absorbed by the in-situ per-anchor calibration. Nevertheless, 
non repeatable or spatially varying distortions remain a limitation of the current 
runtime model and could be addressed in future work through platform specific 
calibration maps or offline FEM/Biot--Savart characterization.
}

\subsection{Measurement Model}
We consider a magnetic positioning setup based on coils acting as source and sensor: a transmitting (TX) coil generates a time varying magnetic field, and a receiving (RX) coil senses its effect. Operation is in the near field at separations that begin at distances $>>$ larger than the coil dimensions, so that the source is compact relative to the TX–RX spacing and the environment is well approximated as linear, homogeneous, and isotropic. 
In this compact source regime, and provided that the orientations of both TX and RX are known (unit normals $\hat{\mathbf{n}}_t$ and $\hat{\mathbf{n}}_r$), we model the TX coil as a magnetic dipole with moment 
\begin{equation}
\mathbf{m}(t)=N_t I(t) A_t \hat{\mathbf{n}}_t
\label{eq:dipole_moment}
\end{equation}
where $t$ is time, $N_t$ is the number of turns of the transmitting coil and $A_t$ is its cross-sectional area. The dipole field in closed form at an observation point $\mathbf{r}$, with $\hat{\mathbf{r}}$ the unit vector from the TX to $\mathbf{r}$ and $r=\|\mathbf{r}\|$, is defined in \Cref{eq:dipole_field_closed}. However, since the system operates under harmonic excitation in the magneto quasistatic regime, and the positioning algorithm processes signal amplitudes (e.g., via FFT extraction), it is possible to henceforth suppress the explicit time dependence $t$. Therefore, the formulations used in the estimation pipeline rely on the phasor magnitudes in \Cref{eq:dipole_model_static}, where $I$ represents the constant current amplitude (peak or RMS).
\begin{equation}
\mathbf{B}(\mathbf{r},t)=\frac{\mu_0}{4\pi r^3}\!\left[3\big(\mathbf{m}(t)\!\cdot\!\hat{\mathbf{r}}\big)\hat{\mathbf{r}}-\mathbf{m}(t)\right]~,
\label{eq:dipole_field_closed}
\end{equation}
\begin{equation}
    \mathbf{m} = N_t I A_t \hat{\mathbf{n}}_t, \quad
    \mathbf{B}(\mathbf{r}) = \frac{\mu_0}{4\pi r^3} \left[ 3(\mathbf{m} \cdot \hat{\mathbf{r}})\hat{\mathbf{r}} - \mathbf{m} \right],
    \label{eq:dipole_model_static}
\end{equation}

Because we use an identical coil on the RX side as magnetic filed sensor\footnote{There exist sensors that directly report $\mathbf{B}$ (e.g., Hall-effect devices), but commercially available options that operate at the frequencies and with the sensitivity relevant to this work are not easily sourced; hence the coil based, voltage sensing approach adopted here.}, it is possible to relate the coupled magnetic field $\mathbf{B}$ at the receiver via the relative measured voltage to the range $r$ and to the relative orientation. 

Since the system operates at a known fixed frequency, the magnitude of the induced electromotive force is directly proportional to the angular frequency $\omega$, the magnetic flux amplitude and the amplifier gain at the RX stage.
The measured voltage amplitude $V_{rx}$ in \Cref{eq:voltage_model} is modeled by explicitly separating the physical induction terms from the electronic amplification, where:
\begin{enumerate*}[label=(\roman*),font=\itshape]
\item $G_{RX}$ represents the total configurable electronic gain of the measurement chain;
\item $\omega = 2\pi f$ is the angular resonance frequency;
\item $N_r$ and $A_r$ are the number of turns and the cross-sectional area of the receiving coil, respectively;
\item $\hat{\mathbf{n}}_r$ is the unit normal vector of the receiving coil.
\end{enumerate*}
\Cref{eq:voltage_model} relates the geometric properties of the field (via the dot product $\mathbf{B} \cdot \hat{\mathbf{n}}_r$) to the scalar value processed by the estimation algorithm.
\begin{equation}
    V_{rx} = \underbrace{G_{RX}}_{\text{RX Gain}} \cdot \underbrace{\left( \omega N_r A_r \right)}_{\text{Transduction Factor}} \cdot \underbrace{\left| \mathbf{B}(\mathbf{r}) \cdot \hat{\mathbf{n}}_r \right|}_{\text{Magnetic Flux}}~.
    \label{eq:voltage_model}
\end{equation}

\subsection{Localization Method}
The localization pipeline runs entirely onboard the nano UAV MCU. It involves three main stages: signal extraction, initial calibration, and iterative position estimation.

\paragraph*{Signal Extraction and Calibration}
The raw signal from the coils is sampled by the ADC and processed via FFT with a flattop window to minimize scalloping loss. 
The amplitude of the peak corresponding to the $i$-th anchor, denoted as $\tilde{V}_{raw, i}$, with $i=1 \ldots 4$, is extracted from the FFT and refined using parabolic interpolation. To map the resulting raw digital values to the physical model in \Cref{eq:voltage_model}, a one-time static calibration is performed before flight during system startup (procedure defined in \Cref{alg:calibration}). During calibration, the UAV is placed at a known reference pose $\mathbf{x}_{ref}$ (the origin of the anchor frame), and the system computes a per-anchor calibration coefficient 
\begin{equation}
    C_i = \tilde{V}_{raw, i}(\mathbf{x}_{ref}) / {V_{model, i}(\mathbf{x}_{ref})}
    \label{eq:calibration_coeff}
\end{equation}

that lumps the electronic gain $G_{RX}$ and the transduction factors, where $V_{model, i}$ is the theoretical voltage predicted by \eqref{eq:voltage_model}. During operation, the calibrated measured voltage is simply obtained as 
\begin{equation}
    V_{meas, i} = \tilde{V}_{raw, i} / C_i
    \label{eq:calibration_formula_runtime}
\end{equation}

.

\paragraph*{Position Estimation via Optimization}
The core of the localization module is a numerical optimizer that solves the inverse magnetic problem: finding the position $\mathbf{x}$ that best explains the observed induced voltages at different bandwidths. 
Unlike trilateration approaches that typically rely on intermediate range estimates, our method directly infers the 3D coordinate $\mathbf{x}$ from the signal amplitudes, thereby preserving the full geometric information contained in the dipole field. 
We formulate this in \Cref{eq:cost_function} as a nonlinear least-squares problem minimizing the residual between the model predictions and the calibrated measurements, where $\Omega$ represents the search space (box constraints) corresponding to the valid flight volume. 
\begin{equation}
    \hat{\mathbf{x}} = \arg\min_{\mathbf{x} \in \Omega} \sum_{i=1}^{N} \left( V_{model, i}(\mathbf{x}, \hat{\mathbf{n}}_r) - V_{meas, i} \right)^2~.
    \label{eq:cost_function}
\end{equation}
We employ the \emph{Nelder-Mead} simplex algorithm for this minimization. This derivative free method is well suited for embedded implementation as it is robust to signal noise and does not require computationally expensive Jacobian calculations. To ensure real time performance (equivalent to an update rate $\approx$ \qty{20}{\hertz}), the optimizer utilizes a warm start strategy: the simplex is initialized around the previous position estimate $\hat{\mathbf{x}}_{k-1}$. Furthermore, the system includes saturation checks: if the signal amplitude of any anchor exceeds the linear range of the ADC (e.g., due to close proximity), that measurement is discarded from the cost function for the current iteration to prevent numerical bias. The complete runtime execution flow is detailed in \Cref{alg:estimation}.

\begin{algorithm}[t]
\caption{Initial Static Calibration}
\label{alg:calibration}
\begin{algorithmic}[1]
\Require Reference pose $\mathbf{x}_{ref}$, Anchor poses $\{\mathbf{p}_i\}$, Number of samples $N_{cal}$
\Ensure Calibration coefficients $\mathcal{C} = \{C_1, \dots, C_4\}$

\State $\mathbf{\bar{v}}_{raw} \gets \mathbf{0}$
\For{$k \gets 1$ to $N_{cal}$} \Comment{Data Accumulation}
    \State $\mathbf{v}_{raw} \gets \textsc{ExtractSignalFeatures}(\text{ADC\_Buffer})$
    \State $\mathbf{\bar{v}}_{raw} \gets \mathbf{\bar{v}}_{raw} + \mathbf{v}_{raw}$
\EndFor
\State $\mathbf{\bar{v}}_{raw} \gets \mathbf{\bar{v}}_{raw} / N_{cal}$ \Comment{Compute Mean Amplitude}

\For{each anchor $i \in \{1 \dots 4\}$}
    \State $V_{model, i} \gets \textsc{DipoleModel}(\mathbf{x}_{ref}, \mathbf{p}_i)$ \Comment{Via Eq. \ref{eq:voltage_model}}
    \State $C_i \gets \bar{v}_{raw, i} / V_{model, i}$ \Comment{Compute Gain}
\EndFor
\State \Return $\mathcal{C}$
\end{algorithmic}
\end{algorithm}

\begin{algorithm}[t]
\caption{Runtime MI Position Estimation Loop}
\label{alg:estimation}
\begin{algorithmic}[1]
\Require Raw\_Signal $\mathcal{S}$, Calibration $\mathcal{C}$, Prev. EKF estimate $\hat{\mathbf{x}}_{k-1}$, Attitude $\hat{\mathbf{n}}_r$
\Ensure Current position estimate $\hat{\mathbf{x}}_k$

\State $\mathbf{v}_{raw} \gets \textsc{SignalProcessing}(\mathcal{S})$
\State $\mathcal{S}_{active} \gets \{1, \dots, 4\}$ \Comment{Set of valid anchors}

\For{each anchor $i$}
    \State $V_{meas, i} \gets v_{raw, i} / C_i$ \Comment{Apply Calibration}
    \If{$V_{meas, i} > V_{sat\_thresh}$} \Comment{Check Saturation}
        \State $\mathcal{S}_{active} \gets \mathcal{S}_{active} \setminus \{i\}$ \Comment{Exclude anchor}
    \EndIf
\EndFor

\State Define cost function (Eq. \eqref{eq:cost_function}) based on active anchors:
\State $J(\mathbf{x}) = \sum_{i \in \mathcal{S}_{active}} (V_{model,i}(\mathbf{x}, \hat{\mathbf{n}}_r) - V_{meas,i})^2$

\State $\mathbf{x}_{opt} \gets \textsc{NelderMead}(J, \text{start}=\hat{\mathbf{x}}_{k-1})$ 

\If{$||\mathbf{x}_{opt} - \hat{\mathbf{x}}_{k-1}|| < \Delta_{outlier}$} \Comment{Outlier Rejection}
    \State $\hat{\mathbf{x}}_k \gets \mathbf{x}_{opt}$
     \State \textsc{send to EKF}($\hat{\mathbf{x}}_k$)
\Else
    \State continue
\EndIf
\State \Return $\hat{\mathbf{x}}_k$
\end{algorithmic}
\end{algorithm}

\subsection{Sensor Fusion with Onboard State Estimation}
\label{subsec:sensor_fusion}

\begin{figure}[H]
    \centering
    \resizebox{\textwidth}{!}{
    \begin{tikzpicture}[
node distance=2.2cm and 1.8cm, 
        auto,
        >=latex,
        thick,
deck/.style={rectangle, draw=black, fill=orange!15, text width=3.5cm, align=center, inner sep=0.3cm, minimum height=2.6cm, rounded corners=5pt, drop shadow, font=\normalsize\bfseries},
        component/.style={rectangle, draw=black!60, fill=white, text width=2.5cm, align=center, rounded corners=2pt, minimum height=1.4cm, font=\small},
        process/.style={rectangle, draw, fill=green!10, text width=2.5cm, align=center, minimum height=1.4cm, drop shadow, font=\small},
        block/.style={rectangle, draw, fill=white, text width=2.5cm, align=center, rounded corners, minimum height=1.4cm, drop shadow, font=\small},
        ekf_block/.style={rectangle, draw, fill=red!10, text width=2.2cm, align=center, rounded corners, minimum height=1.4cm, drop shadow, font=\small},
        image_node/.style={inner sep=0pt, align=center},
        line_label/.style={font=\small, align=center, auto, fill=white, inner sep=2pt, text=black!90},
        platform/.style={draw, dashed, inner sep=0.5cm, rounded corners, fill=gray!5, label={[anchor=south, yshift=-0.6cm, font=\normalsize\bfseries]south:#1}}
    ]

\node [deck] (anchor_deck) {AnchorDeck};

\node [block, fill=yellow!20, below=0.3cm of anchor_deck] (battery) {\textbf{LiPo} \small 12 Watt};

\coordinate (coils_start_x) at ($(anchor_deck.east) + (3.5cm, 0)$);
        
\node [image_node, label={[font=\small, yshift=0.2cm]west:$f_1$}] (coil1) at ($(coils_start_x) + (0, 1.2)$) {\includegraphics[width=1.0cm]{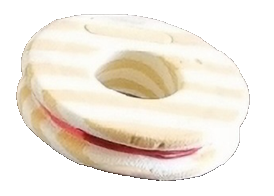}};
        \node [image_node, label={[font=\small, yshift=0.25cm]west:$f_2$}] (coil2) at ($(coils_start_x) + (0, 0.4)$) {\includegraphics[width=1.0cm]{fig_coil.png}};
        \node [image_node, label={[font=\small, yshift=0.3cm]west:$f_3$}] (coil3) at ($(coils_start_x) + (0, -0.4)$) {\includegraphics[width=1.0cm]{fig_coil.png}};
        \node [image_node, label={[font=\small, yshift=0.4cm]west:$f_4$}] (coil4) at ($(coils_start_x) + (0, -1.2)$) {\includegraphics[width=1.0cm]{fig_coil.png}};
        
        \node [below=0.0cm of coil4, font=\small\bfseries] {4x Anchors};

\begin{scope}[on background layer]
            \node[platform={UGV Platform (Unitree A1)}, fit=(anchor_deck) (coil1) (coil4) (battery)] {};
        \end{scope}

\node [right=1.2cm of coil2, yshift=0.3cm, align=center, font=\itshape\normalsize] (channel) {Dipole Field\\$\vec{B}_{tot} = \sum \vec{B}_i$};

\node [image_node, right=1.2cm of channel, label={[font=\small, align=center]below:\textbf{RX Coil}\\(Tag)}] (rx_coil) {\includegraphics[width=1.0cm]{fig_coil.png}};

\node [block, fill=blue!10, right=1.8cm of rx_coil] (mag_deck) {\textbf{MagneticDeck}\\ \small Gain 10-1000x};
        \node [process, right=1.8cm of mag_deck] (dsp) {\textbf{Signal Extr.}\\ \small ADC + FFT};
        \node [process, right=1.8cm of dsp] (solver) {\textbf{Estimator}\\ \small Nelder-Mead};
        \node [ekf_block, right=1.8cm of solver] (ekf) {\textbf{EKF}\\ \small Fusion};
        \node [right=1.2cm of ekf, font=\large\bfseries, align=center] (state) {State\\$(x,y,z,\psi)$};

\node [block, fill=gray!10, below=1.5cm of solver] (imu) {\textbf{IMU / Flow}\\ \small Inertial \& Optical};
        \node [block, fill=cyan!10, below=1.5cm of ekf] (uwb) {\textbf{Loco UWB}\\ \small TWR/TDOA};

\begin{scope}[on background layer]
            \node[platform={nano UAV (Crazyflie)}, fit=(rx_coil) (mag_deck) (state) (uwb) (imu)] {};
        \end{scope}

\draw [->, red!70!black, line width=1.2pt] (battery.north) -- (anchor_deck.south);

\draw [->] ([yshift=1.2cm]anchor_deck.east) -- (coil1.west); 
        \draw [->] ([yshift=0.4cm]anchor_deck.east) -- (coil2.west);
        \draw [->] ([yshift=-0.4cm]anchor_deck.east) -- (coil3.west);
        \draw [->] ([yshift=-1.2cm]anchor_deck.east) -- (coil4.west);
        
\node [line_label, align=center] at ($(anchor_deck.east)!0.5!(coil1.west) + (0, 0.9)$) {AC Signals};     

\draw [->, dashed, blue] (coil1.east) -- (channel);
        \draw [->, dashed, blue] (coil2.east) -- (channel);
        \draw [->, dashed, blue] (coil3.east) -- (channel);
        \draw [->, dashed, blue] (coil4.east) -- (channel);
        \draw [->, dashed, blue] (channel) -- (rx_coil);

\draw [->] (rx_coil) -- node[line_label] {Induced V} (mag_deck);
        \draw [->] (mag_deck) -- node[line_label] {Analog Sig} (dsp);
        \draw [->] (dsp) -- node[line_label] {Amplitudes} (solver);
        \draw [->] (solver) -- node[line_label] {Pos $\hat{x}$} (ekf);
        \draw [->] (ekf) -- (state);
        
\draw [->] (imu.north) -- ++(0, 0.5) -| node[line_label, near start] {Odometry} ([xshift=-0.5cm]ekf.south);
        \draw [->] ([xshift=0.5cm]uwb.north) -- node[line_label, swap] {Range} ([xshift=0.5cm]ekf.south);

    \end{tikzpicture}
    }
    \caption{\textbf{System Overview Block Diagram.} The \textit{AnchorDeck} (orange box) houses the signal generation and driving logic, powered by the UGV battery. It directly drives four independent anchor coils. The magnetic field is received by the UAV, processed alongside UWB and IMU data within the EKF for state estimation.}
    \label{fig:overview_sistema_final_v5}
\end{figure}

The absolute position estimate provided by the magnetic localization pipeline is fused with the UAV's high-rate onboard sensors to obtain a robust state estimate for feedback control. We utilize the nano UAV stock EKF, which runs onboard the MCU. Notably, the UWB localization is natively supported, with the fusion architecture  depicted in \Cref{fig:overview_sistema_final_v5}.

The position estimate $\hat{\mathbf{x}}_k$ output by the Nelder-Mead optimizer in \Cref{alg:estimation} is treated as an asynchronous absolute position measurement update for the EKF. The measurement noise covariance matrix $\mathbf{R}_{mag}$ associated with these updates is tuned based on the static characterization of the system; specifically, the standard deviation parameters are set according to the values determined in  \cite{brunacci2023fusion,DeAngelis2017}. This direct injection of position allows the EKF to correct the drift inherent in the inertial integration and optical flow, effectively pinning the drone's frame of reference to the UGV frame $\{\mathcal{B}\}$.

While the magnetic system provides a 3D position estimate, the vertical component ($z$) is often the least observable axis in planar anchor configurations and can be noisy. To ensure robust altitude hold, we augment the system with a downward-facing Time-of-Flight (ToF) laser rangefinder (the VL53L1x series for STMicroelectronics). However, the integration of the ToF sensor presents a specific challenge in our heterogeneous setup: the UAV takes off from the back of the UGV, and as the drone performs a lateral maneuver to exit the UGV's footprint, the ground reference for the ToF sensor changes abruptly from the UGV's back to the floor, causing a discontinuity (step) in the raw distance measurement. If fed directly to the EKF, this jump would be interpreted as a sudden gain in altitude, leading to destabilizing control reactions. To address this, we implemented a custom filtering logic (state machine) within the sensor driver. The algorithm monitors the numerical derivative of the raw distance measurement $\dot{d} \approx (d_k - d_{k-1})/\Delta t$. The logic operates as follows:
\begin{enumerate}[label=(\roman*),font=\itshape]
\item Nominal Flight: When the derivative is below a safety threshold ($\dot{d} < \delta_{thresh}$), the sensor is assumed to be tracking a continuous surface. The measurement is passed to the estimator with a standard compensation offset.
\item Discontinuity Detection (The "Step"): When the drone flies off the UGV, a large spike in the derivative is detected ($\dot{d} > \delta_{thresh}$). The filter identifies this as a surface change rather than vertical motion.
\item Output Smoothing: Upon detection of the step, the filter holds the previous valid compensated altitude value, effectively ignoring the jump in the raw data, while internally realigning the reference baseline $d_{t0}$.
\end{enumerate}
This strategy allows the EKF to maintain a continuous estimate of the altitude relative to the takeoff frame, seamlessly transitioning from "relative-to-robot" to "relative-to-ground" measurements without requiring external triggers.

\section{System Architecture}
\label{sec:system}

\begin{figure}[H]
    \centering
    
\begin{subfigure}[t]{0.99\linewidth}
        \centering
        \includegraphics[width=\linewidth,height=0.26\textheight,keepaspectratio]{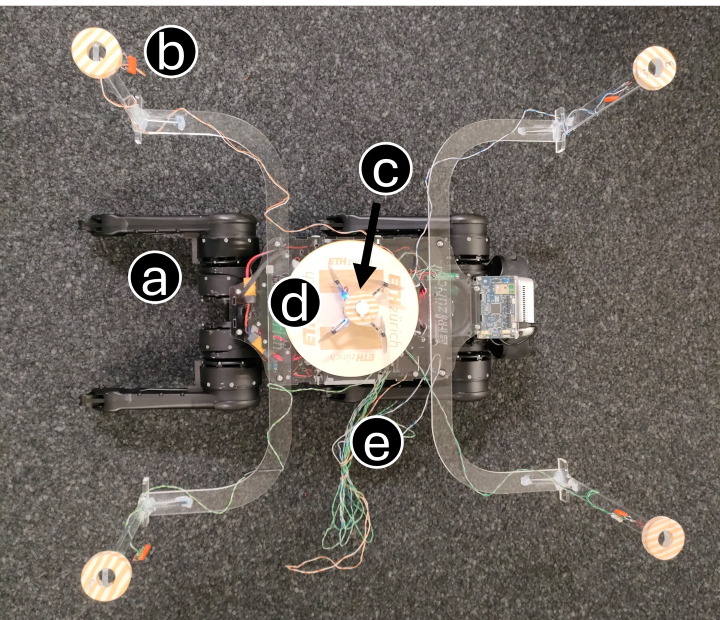}
    \end{subfigure}
    
    \vspace{1mm} 

\begin{subfigure}[b]{0.99\linewidth}
        \centering
        \includegraphics[width=\linewidth,height=0.26\textheight,keepaspectratio]{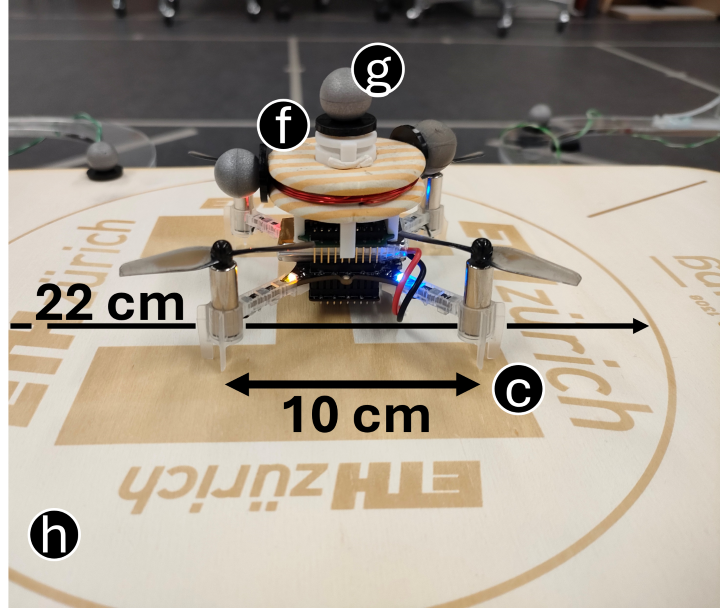}
    \end{subfigure}
    
    \caption{\textbf{Custom Magnetic Hardware.} (a) The Unitree A1 legged robot UGV. (b) One of the four lightweight MI coils used for precise localization. (c) The Crazyflie 2.1 nano UAV. (d) The landing pad with a diameter of \qty{22}{\cm} mounted on UGV. (e) The wiring for the AnchorDeck and the four coils, plus the connections for the wireless battery charger. (f) The ultra-lightweight MI coil mounted on the Crazyflie 2.1 nano UAV and the MagneticDeck. (g) Motion caption system markers used to assess the system performances.}
    \label{fig:hardware_composite}
\end{figure}

In this work, we investigate heterogeneous cooperation and teaming of two markedly different robotic platforms—a legged robot and a nano UAV. These robots differ substantially in onboard computation, payload capacity, sensing power budget, and physical size, which severely constrains the class of localization sensors that each platform can support. Moreover, to achieve practical and robust cooperation in real environments, the system must remain independent of any external infrastructure for localization, environmental sensing, and computation. The system described in this section has been designed explicitly around these requirements. For the scope of this paper, we selected two commercial platforms extendable with custom hardware and software.

The ground platform consists of a Unitree~A1 quadrupedal robot equipped with a custom lightweight backpack for onboard sensing and computation. The robot has a nominal footprint of approximately $\qtyproduct{500 x 300 x 400}{\milli\meter}$ and a mass of $\qty{12.0}{\kilo\gram}$ in its base configuration, a battery of $\qty{91}{\watt\per\hour}$, and is equipped with twelve torque-controlled joints and onboard proprioceptive sensing in the form of joint encoders and a 9-DoF IMU. The A1 supports payloads of up to $\qty{5}{\kilo\gram}$, of which $\qty{3.15}{\kilo\gram}$ are used by the backpack, including an additional $\qty{96}{\watt\hour}$ battery and an Intel~NUC used for navigation and high-level processing. The backpack also serves as a rigid mounting structure for the localization hardware and the landing deck introduced in \cref{fig:hardware_composite} and detailed below.

The Crazyflie 2.1 is a palm-sized (\qtyproduct{9.2 x 9.2}{\cm}), open-source nano-quadrotor platform designed for research and education in aerial robotics. It features a modular architecture with expansion decks that enable rapid prototyping of sensors, actuators, and communication systems. The onboard STM32F4 microcontroller handles flight control and sensor fusion. Equipped with a 9-DoF IMU, we extended its onboard sensors with the \emph{Flow deck v2} from Bitcraze. The total power budget available for sensing and computation is $\sim$\qty{1}{\watt} \cite{niculescu2025ultra}. We selected the Crazyflie 2.1 because it offers a unique combination of low weight, extensibility, and real time control, making it ideal for experiments where SWaP constraints are critical.

\Cref{fig:hardware_composite} outlines the overall system architecture. The UGV is equipped with a landing pad with a size of \qtyproduct{25 x 44}{\cm} made with transparent plexiglas (PMMA) and Polywood. The Crazyflie recharging coil (based on the \emph{Qi 1.2 charger deck} from Bitcraze) is positioned in the marked area, with a \text{\diameter}\qty{22}{\cm}, which the UAV has to center during landing. Therefore, to center the nano UAV inside the landing area, the relative landing precision needs to provide a consistent accuracy $<$\qty{10}{\cm}. Four transmitting coils are mounted on rigid supports attached to the quadruped’s backpack (\Cref{fig:hardware_composite}), positioned at each corner with a \text{\diameter}\qty{2}{\centi\meter} plexiglas pipe \qty{25}{\cm} tall. On the other side, a single receiving coil is mounted on the Crazyflie using a custom 3D-printed holder. These constitute the key hardware elements of the magnetic anchor-tag system: the four coils on the legged robot act as anchors, and the receiving coil on the UAV acts as the tag.  The coils, shown in detail in \Cref{fig:hardware_composite}, were hand-wound around lightweight, non-ferromagnetic supports. The total weight of the Crazyflie is \qty{47}{\gram}, of which \qty{9}{\gram} are added by the coil and the holder. The receiving coil mounted on the Crazyflie connects to a dedicated conditioning board, the MagneticDeck (\Cref{fig:hardware_composite}). The deck provides solder pads for the coil terminals and includes a parallel capacitor to establish an initial resonant gain. Beyond resonance, the MagneticDeck includes two amplification stages:
\begin{enumerate*}[label=(\roman*),font=\itshape]
    \item a fixed-gain $10\times$ amplifier, and
    \item a digitally programmable amplifier up to $100\times$ dynamically controlled by the onboard MCU.
\end{enumerate*}
The final stage of the amplification and conditioning system is connected to the MCU internal 12-bit ADC, sampling at \qty{518}{\kilo\hertz}.
The received voltage is a modulation of the four anchor signals; the amplitude and main frequency components of each must be extracted in the frequency domain, as described in \Cref{sec:Receiving side}.

All coils share identical parameters (the number of turns, wire type, and diameter) except for unavoidable manufacturing tolerances, ensuring consistent magnetic behavior. Specifically, each coil has a radius of \qty{1.9}{\cm} and 5 turns. Each anchor coil on the quadruped generates an AC magnetic dipole field by driving a sinusoidal current at a unique frequency (210, 199, 189, and \qty{181}{\kilo\hertz}). Conversely, the tag coil on the UAV behaves as a passive sensor: immersed in the superposition of the four dipole fields, it generates an electromotive force proportional to the received magnetic flux.

Two independent body frames and coordinate are present throughout the system, which need to be aligned: one relative to the quadruped’s body (and moving with it), and one attached to the UAV. The Crazyflie initializes its own frame at startup using onboard inertial sensors; due to drift, the magnetic localization system subsequently provides the primary absolute reference, effectively tying the UAV frame to the quadruped frame.

{\color{orange}\color{black}

\begin{table}[htbp] 
\color{orange}\color{black} 
\centering
\footnotesize 
\linespread{1}\selectfont 
\renewcommand{\arraystretch}{1.1} 
\caption{Main parameters of the MI anchor-tag localization system.}
\label{tab:system_parameters}
\begin{tabularx}{\linewidth}{@{} >{\raggedright\arraybackslash\hsize=1.1\hsize}X >{\raggedright\arraybackslash\hsize=0.9\hsize}X @{}}
\toprule
\textbf{Parameter} & \textbf{Value} \\
\midrule

\multicolumn{2}{@{}l}{\textbf{Platform}} \\
\quad UGV / UAV                 & Unitree A1 / Crazyflie 2.1 \\
\quad UAV mass / MI payload     & 47 g / 9 g \\
\quad Landing pad / Target      & 25$\times$44 cm / $\varnothing$ 22 cm \\
\addlinespace

\multicolumn{2}{@{}l}{\textbf{Coils}} \\
\quad TX anchors / RX tag       & 4 / 1 passive coil \\
\quad Radius / Turns            & 1.9 cm / 5 \\
\quad TX frequencies            & 181, 189, 210, 199 kHz \\
\quad TX model current          & 0.5 A \\
\quad TX dipole axes $u_i^B$    & $[0, 0, 1]^\top$ \\
\quad Anchor coordinates $p_i^B$& $(\pm 0.295, \pm 0.250, 0.250)$ m \\
\addlinespace

\multicolumn{2}{@{}l}{\textbf{Signal Acquisition \& FFT}} \\
\quad ADC resolution / Rate     & 12 bit / 518.5 kHz \\
\quad ADC range                 & 0--3.0 V \\
\quad Analog gain (used/range)  & 100$\times$ / 10--1000$\times$ \\
\quad FFT length / Overlap      & 2048 / 0\% \\
\quad Bin resolution            & 253 Hz \\
\quad Anchor FFT bins           & 715, 747, 830, 786 \\
\quad Window / Peak refinement  & Flattop / Parabolic interpolation \\
\quad Saturation threshold      & 1.2 V \\
\addlinespace

\multicolumn{2}{@{}l}{\textbf{Estimator}} \\
\quad Calibration samples       & 500 for each anchor\\
\quad Calibration pose          & $(0, 0, 0.035)$ m - Landing pad center \\
\quad Solver                    & warm started Nelder--Mead \\
\quad Active anchors            & 4 (fallback to 3 if saturated) \\
\quad Simplex size              & $(0.05, 0.05, 0.05)$ m \\
\quad Max iterations            & 20 \\
\quad Tolerances                & $10^{-6}$ (cost), $10^{-5}$ m (position) \\
\quad EKF position std. dev.    & 0.06 m \\
\quad XY outlier threshold      & 0.50 m \\
\quad Firmware rate             & 30 Hz \\

\bottomrule
\end{tabularx}
\end{table}

To make the system reproducible and to clarify which quantities are used online by the embedded estimator, Table~\ref{tab:system_parameters} summarizes the hardware, signal-processing, estimation, and integration parameters used in the experimental campaign. Values marked as configuration parameters are directly taken from the firmware or mechanical design, while quantities not logged during the original flight campaign are explicitly reported as not characterized rather than inferred.
}

\subsection{Magnetic Deck on the nano UAV}
\label{sec:Receiving side}
\Cref{fig:overview_sistema_final_v5} shows the full processing chain. After amplification, the signal is digitized by the Crazyflie’s onboard ADC. A Flattop window is applied to increase amplitude accuracy, and a real time FFT isolates the four frequency components, each corresponding to a different anchor. Peak amplitude estimates are refined via standard parabolic interpolation around the FFT bin associated with the anchor’s excitation frequency. For anchor $k$, the sinusoid at frequency $f_k$ appears at FFT bin $b_k$  with interpolated amplitude $A_k$. These amplitudes are forwarded to the magnetic inverse dipole localization module. The estimated position is then fused in the Crazyflie’s EKF along with inertial, UWB, and optical-flow data.  
\section{Experimental Evaluation}
\label{sec:experimental_evaluation}
Experiments were conducted in an indoor flight arena equipped with a multi camera motion capture system (Vicon), utilized \emph{exclusively} to provide ground truth (GT) for validation. We employed the heterogeneous system architecture detailed in \Cref{sec:system}.

The system evaluation, which includes RMSE and landing success rate, relies on the coordinate frames $\{B\}$ (Anchor/UGV Body) and $\{U\}$ (Tag/UAV Body) defined in \Cref{sec:mag_model}. For this reason, motion capture markers were placed on both robots to track their poses in the world frame $\{W\}$, as visible in \Cref{fig:hardware_composite}. For quantitative assessment, all quantities (including motion capture GT and the nano UAV on-board EKF estimate driven by the magnetic localization) are expressed in the world frame $\{W\}$ by applying the corresponding rigid body transforms (e.g., $^{W}\!T_{B}$) at each timestep.

\subsection{Field Experimental Protocol}
\label{sec:scenarios}
\begin{figure}[H]
    \centering
\begin{subfigure}[b]{0.32\textwidth}
        \centering
        \includegraphics[width=\linewidth]{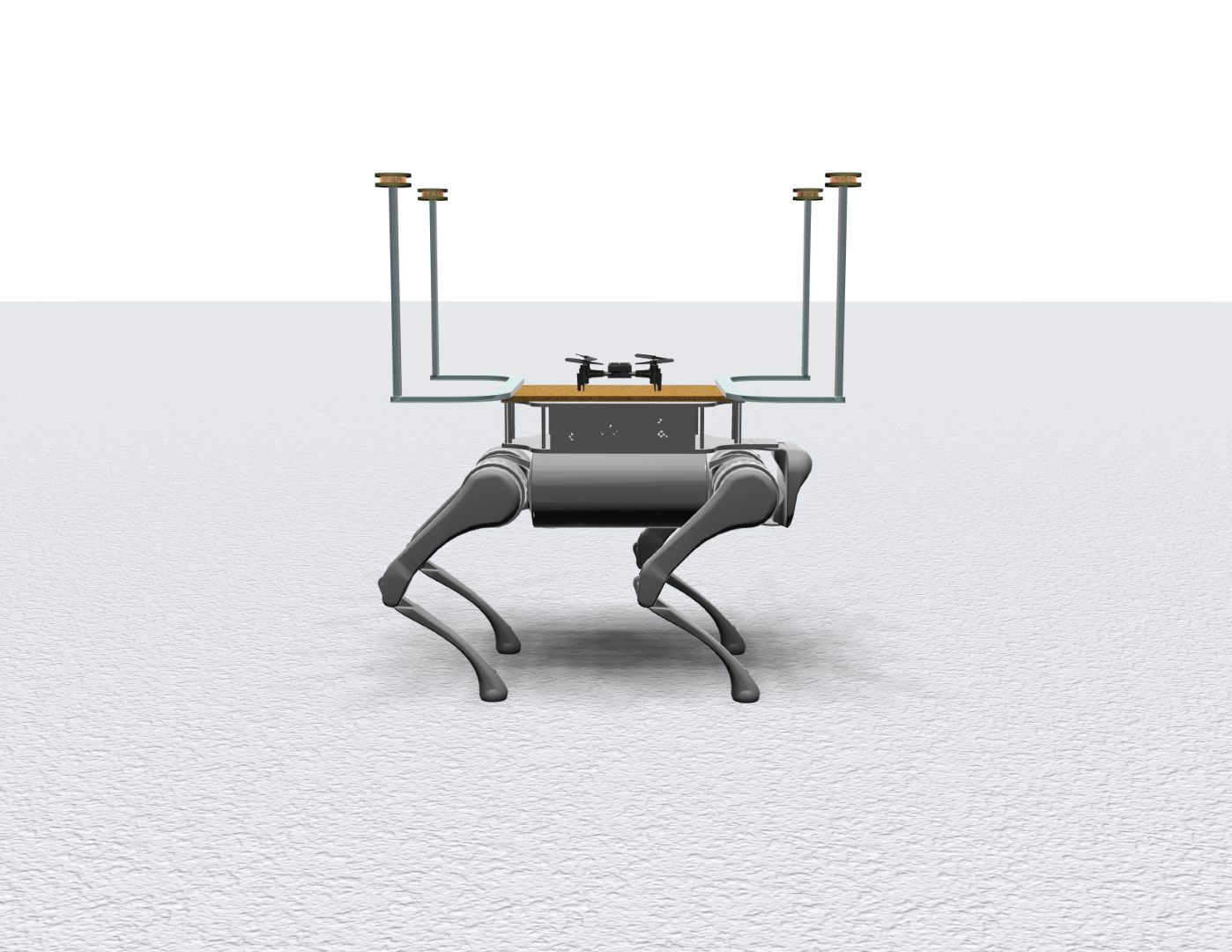}
        \caption{S1 - Take-off}
        \label{fig:s1_takeoff}
    \end{subfigure}
    \hfill
    \begin{subfigure}[b]{0.32\textwidth}
        \centering
        \includegraphics[width=\linewidth]{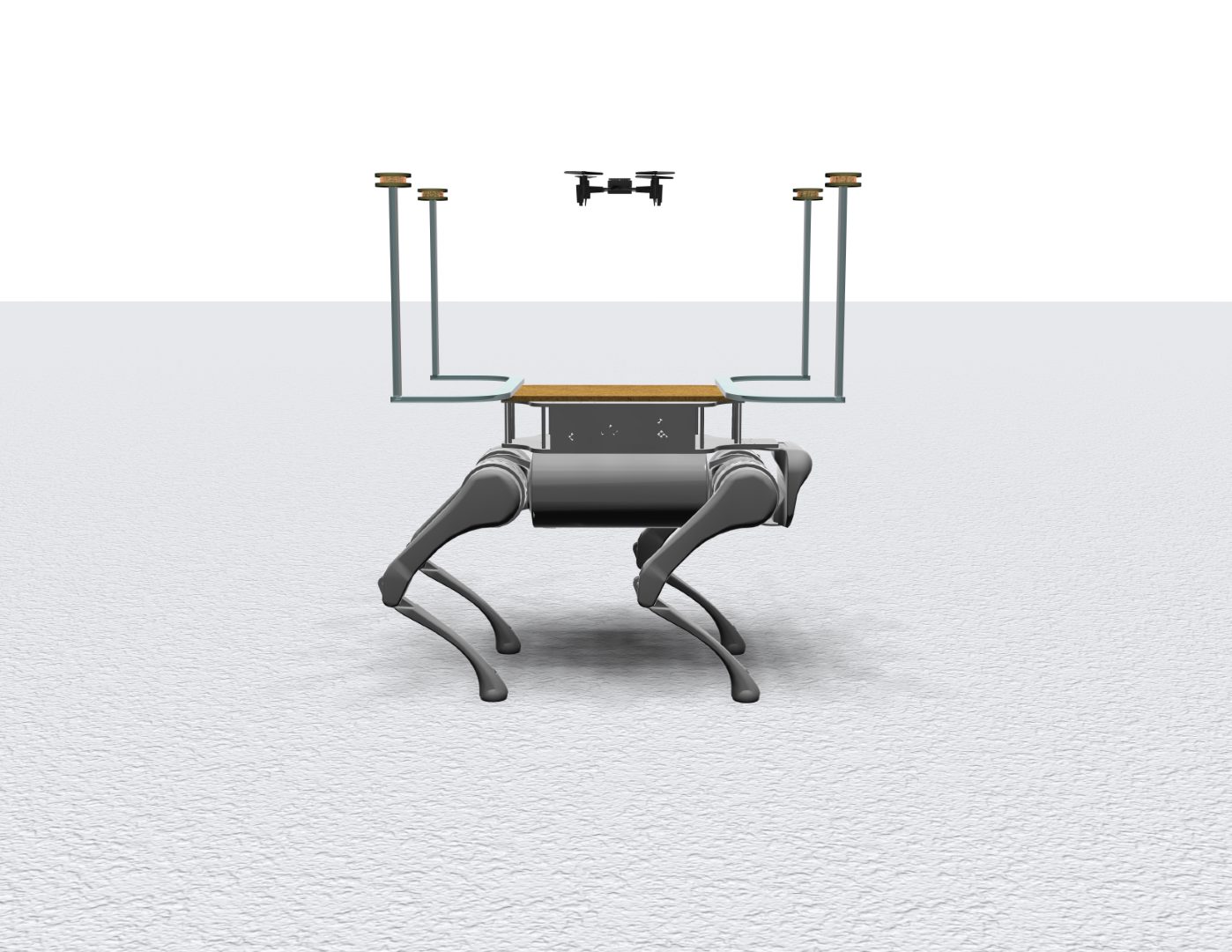}
        \caption{S1 - Hovering}
        \label{fig:s1_hover}
    \end{subfigure}
    \hfill
    \begin{subfigure}[b]{0.32\textwidth}
        \centering
        \includegraphics[width=\linewidth]{fig_s1_takeoff_landing.jpg}
        \caption{S1 - Landing}
        \label{fig:s1_landing}
    \end{subfigure}
    
\par\vspace{-0.2cm} \noindent\rule{\linewidth}{0.4pt} \vspace{0.05cm} 

\begin{subfigure}[b]{0.32\textwidth}
        \centering
        \includegraphics[width=\linewidth]{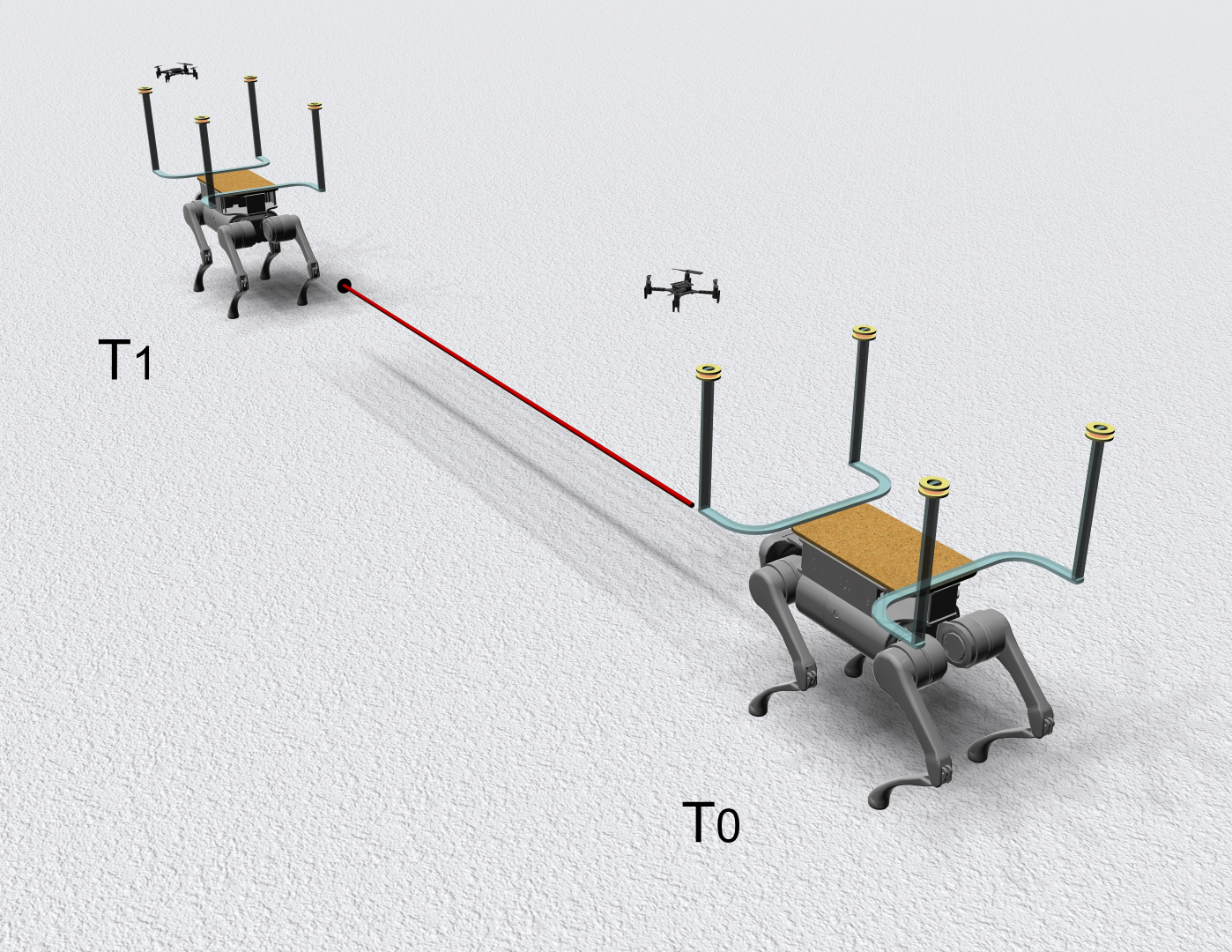}
        \caption{S2 - 3D View}
        \label{fig:s2_3d}
    \end{subfigure}
    \hfill
    \begin{subfigure}[b]{0.32\textwidth}
        \centering
        \includegraphics[width=\linewidth]{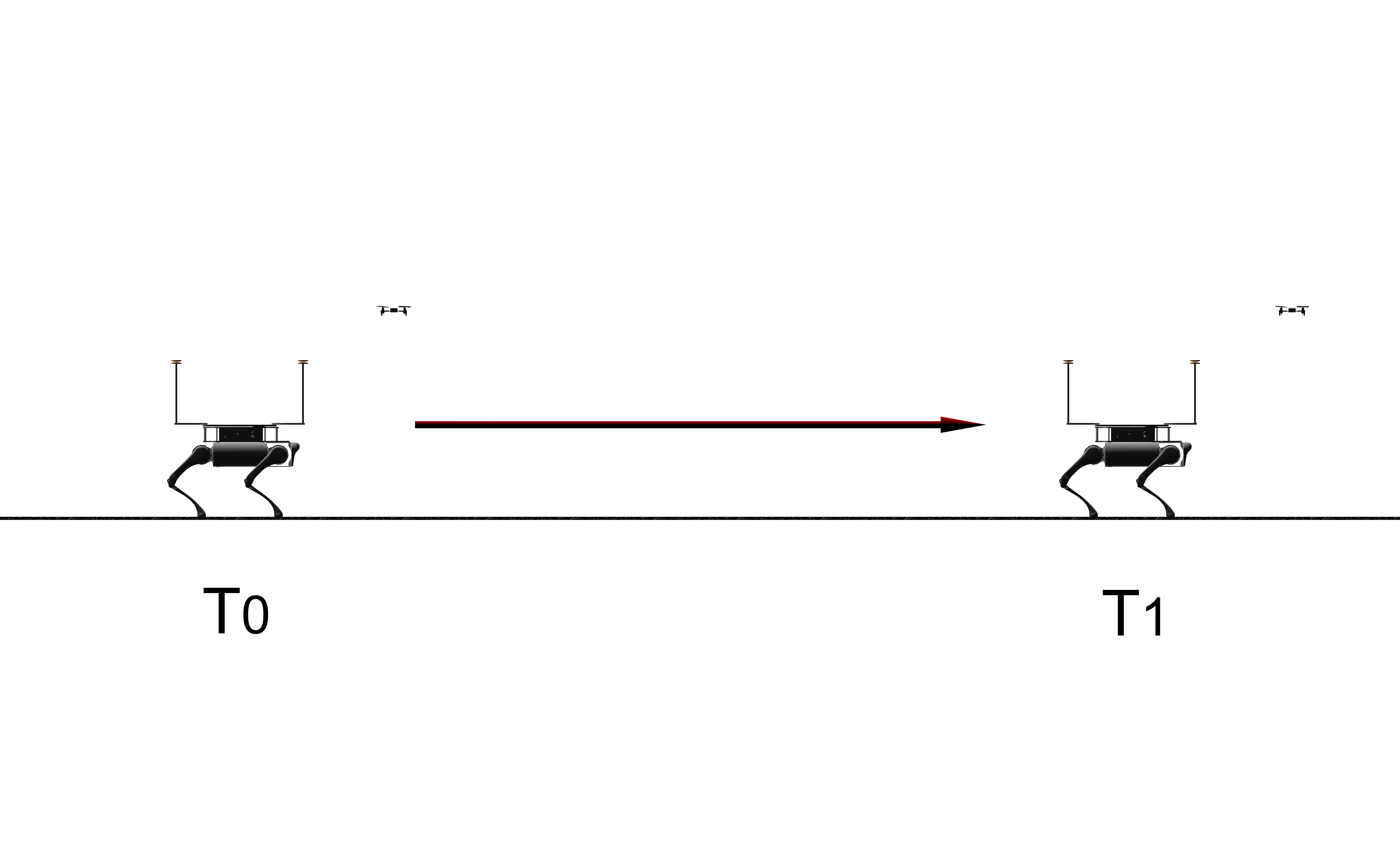}
        \caption{S2 - Lateral View}
        \label{fig:s2_side}
    \end{subfigure}
    \hfill
    \begin{subfigure}[b]{0.32\textwidth}
        \centering
        \includegraphics[width=\linewidth]{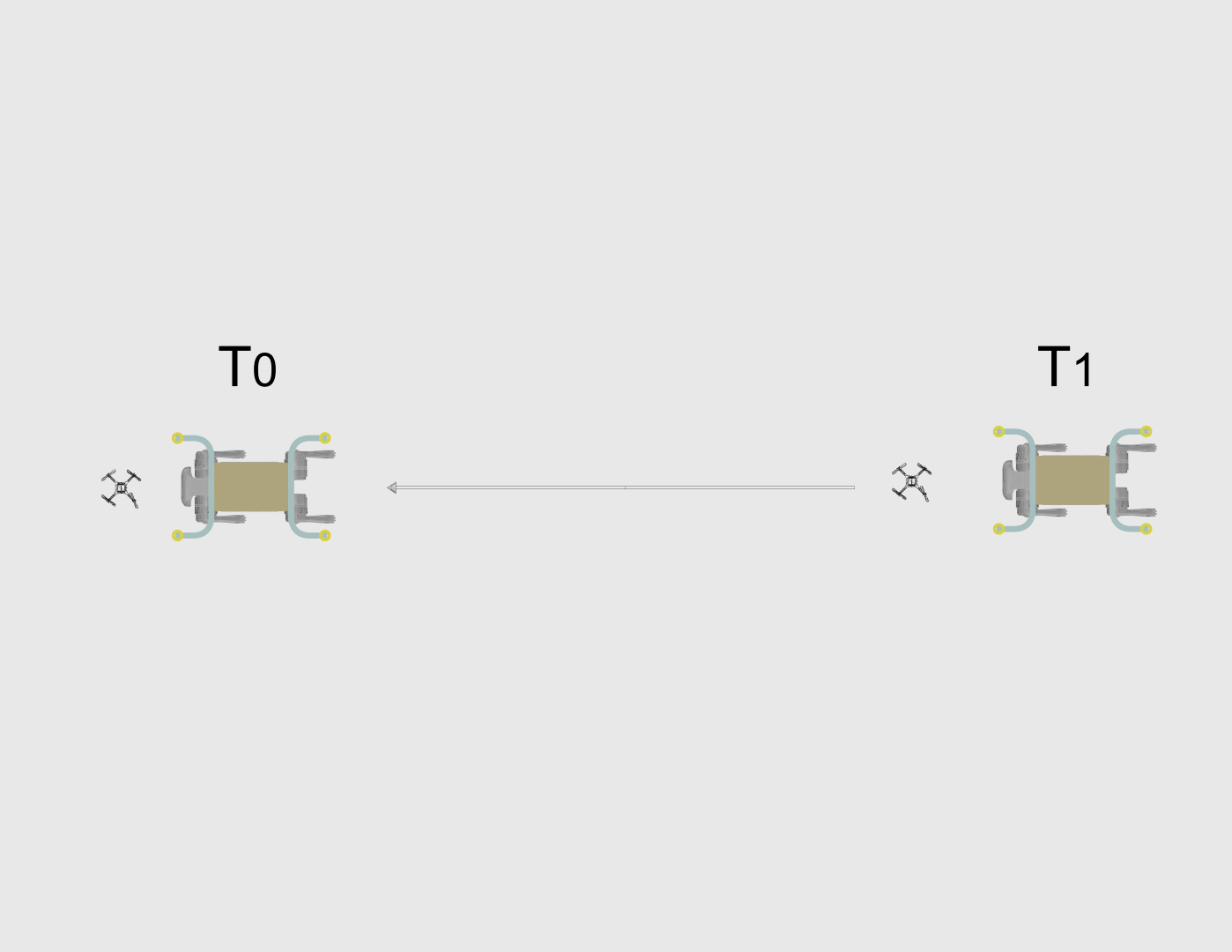}
        \caption{S2 - Top View}
        \label{fig:s2_top}
    \end{subfigure}
    
\par\vspace{-0.2cm} \noindent\rule{\linewidth}{0.4pt} \vspace{0.05cm} 

\begin{subfigure}[b]{0.48\textwidth}
        \centering
        \includegraphics[width=\linewidth]{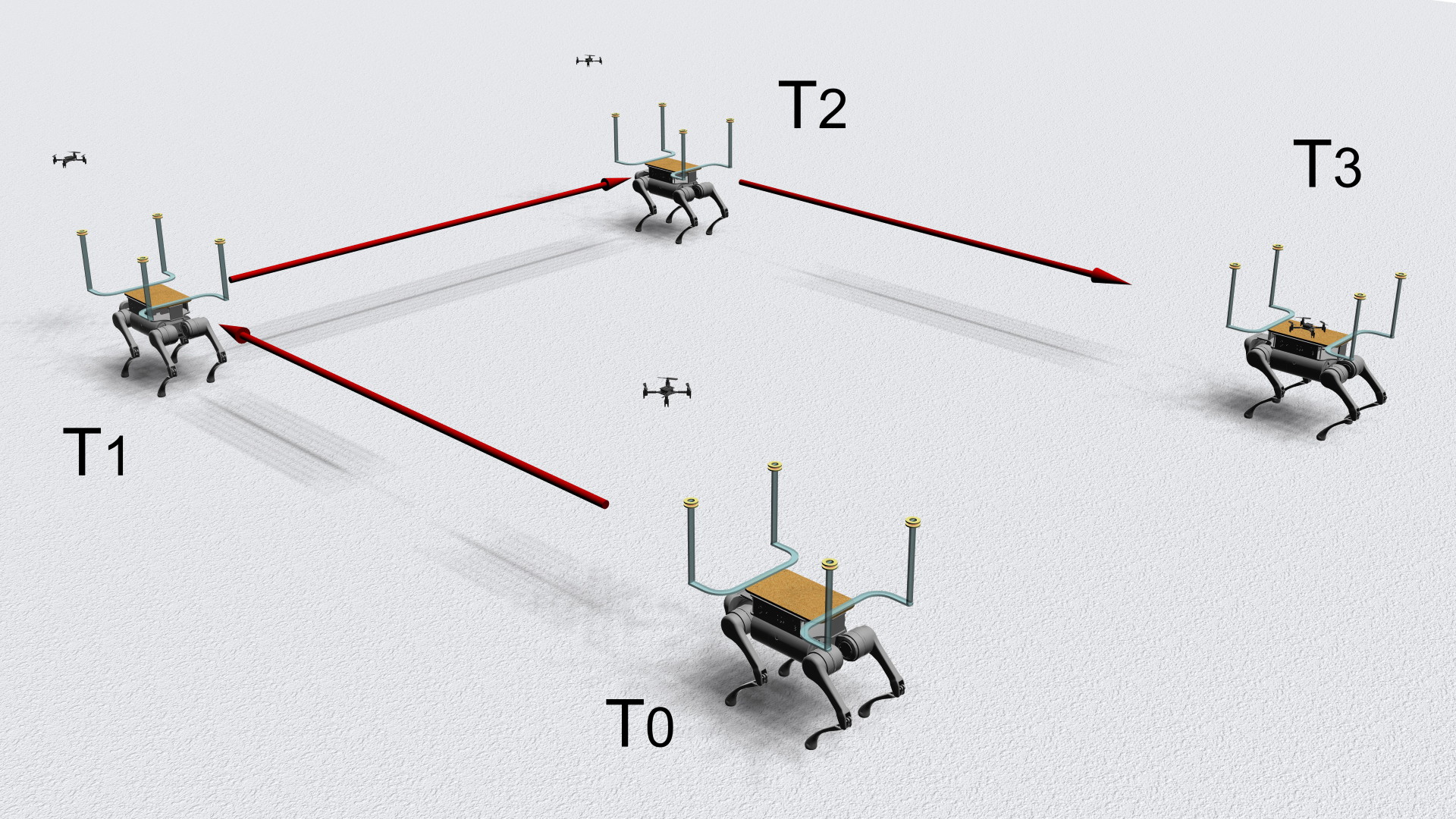}
        \caption{S3 - Composite 3D Motion}
        \label{fig:s3_view1}
    \end{subfigure}
    \hfill
    \begin{subfigure}[b]{0.34\textwidth}
        \centering
        \includegraphics[width=\linewidth]{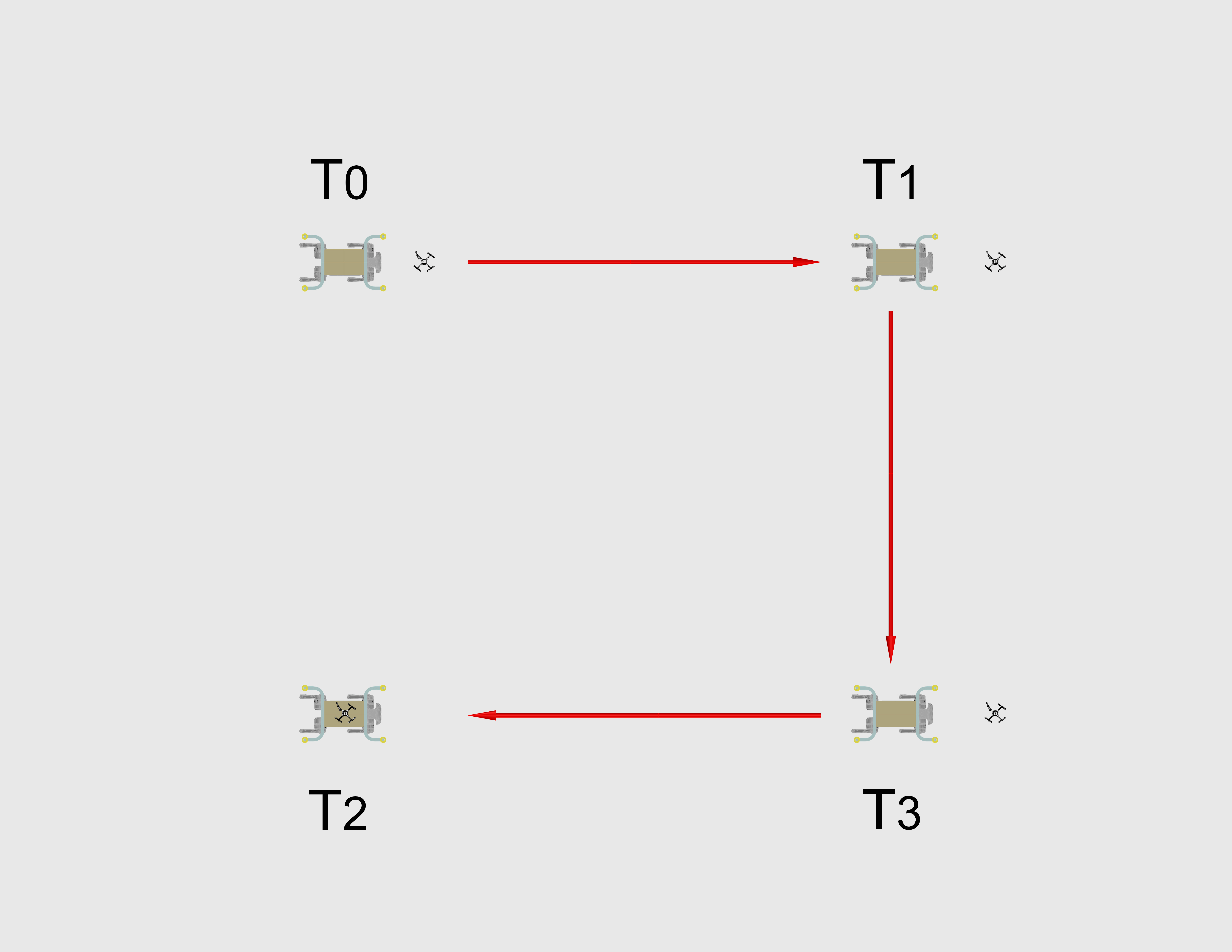}
        \caption{S3 - Composite 2D Motion}
        \label{fig:s3_view2}
    \end{subfigure}
    
    \vspace{-2mm}
    \caption{Visual overview of the experimental validation scenarios using the proposed magnetic localization system. 
    \textbf{(a)-(c) S1 (Static Hovering \& Landing):} The nano UAV performs an autonomous sequence of takeoff, hovering, and precision landing on the stationary UGV.
    \textbf{(d)-(f) S2 (Linear Tracking \& Landing):} The nano UAV tracks the UGV moving along a linear trajectory. Visualization includes (d) 3D perspective, (e) lateral view showing the relative distance maintenance ($T_0$ to $T_1$), and (f) top-down view of the alignment.
    \textbf{(g)-(h) S3 (Composite Motion Tracking):} The UAV tracks and follows the UGV performing complex planar maneuvers with varying velocity and direction, maintaining the relative position within the magnetic workspace.}
    \label{fig:experimental_scenarios}
\end{figure}

The experimental protocol is depicted in \Cref{fig:experimental_scenarios} and defined as follows: upon the first system startup, the initial static calibration \Cref{alg:calibration} is executed to normalize per-anchor gains. It requires few seconds during which the overall system is static. After calibration, the nano UAV autonomously takes off and executes the mission; for subsequent missions following a landing, the calibration is not repeated and the previously estimated gains are used. 
{\color{orange}\color{black}
For the scope of this paper, we isolate and evaluate the contribution of the proposed MI localization layer. The experimental baseline is the native infrastructure-less Crazyflie estimator based on IMU, optical flow, and altitude sensing, with magnetic aiding disabled. We do not evaluate UWB as a direct experimental baseline. UWB and other RF localization systems are considered complementary mid-range acquisition modalities, whereas the focus of this work is the final close range relative localization and docking phase below the magnetic system's operational range. In this view, we consider UWB localization as a valid solution for long range missions (UAV-UGV distance $>$\qty{1}{\meter}), for nano UAV since is already been proven feasibile within the SWaP constraint of these platforms \cite{polonelli2022performance,schindler2023relative}.}

We evaluate the system across three scenarios and a comparative baseline:
\begin{enumerate*}[label=(\roman*),font=\itshape]
\item \textbf{Baseline} (Flow-only): the nano UAV holds position and track the UGV using only its native sensing stack (IMU + optical flow), with magnetic aiding disabled. This serves as a baseline to benchmark the drift accumulation inherent to infrastructure-less nano UAVs in comparison with our MI positioning.
\item \textbf{S1} (Static Hovering \& Landing): the UGV remains stationary. The nano UAV stabilizes at a setpoint $\mathbf{x}_{ref}$ above the docking pad using magnetic localization, then performs an autonomous precision landing. This test aims to measure the drift accumulation and positioning noise over time. The movement sequence is shown in \Cref{fig:s1_takeoff}, \Cref{fig:s1_hover}, \Cref{fig:s1_landing}.
\item \textbf{S2} (Linear Tracking \& Landing): the UGV executes bounded planar motions (forward--backward). The nano UAV tracks the moving reference frame $\{B\}$ and attempts autonomous landing while the landing pad is in motion. The movement sequence is shown in \Cref{fig:s2_3d}.
\item \textbf{S3} (Composite Motion Tracking): this tests simulate a UGV-UAV shared mission, where the nano UAV works as flying sensor for the ground legged robot. The UGV follows complex piecewise-smooth trajectories involving changes in speed and direction, while the nano UAV is commanded to maintain a fixed relative hover position. The movement sequence is shown in \Cref{fig:s3_view1}.
\end{enumerate*}

\subsection{Evaluation Metrics}
\label{subsec:metrics}
The system performance for relative tracking and landing accuracy are evaluate through the position RMSE and the success rate. To quantify tracking accuracy, we compute the 3D RMSE in \Cref{eq:rmse} between the UAV position and the GT over $N$ samples, where $\mathbf{p}^{\,\mathrm{est}}_{B}$ is the onboard position estimate and $\mathbf{p}^{\,\mathrm{GT}}_{B}$ is the ground truth. We also report axis-wise RMSE ($x, y, z$) to analyze anisotropy.
\begin{equation}
\mathrm{RMSE}_{3D} \;=\; \sqrt{\frac{1}{N}\sum_{k=1}^{N}\bigl\| \mathbf{p}^{\,\mathrm{est}}_{B}(k) - \mathbf{p}^{\,\mathrm{GT}}_{B}(k) \bigr\|^{2}}~.
\label{eq:rmse}
\end{equation}

Regarding the success rate, a trial is considered successful if the UAV completes the task (hovering or landing or both) without strictly violating safety bounds. 
We declare \emph{failure} if: \begin{enumerate*}[label=(\roman*),font=\itshape] \item the instantaneous position error exceeds \qty{0.5}{\meter} (safe volume violation), or \item  the controller aborts due to signal loss or estimator divergence.
\end{enumerate*}
 
\section{Experimental Results}
\label{sec:results}

We validate the proposed infrastructure-less MI localization system through extensive real world  experiments. The evaluation focuses on three key aspects: \begin{enumerate*}[label=(\roman*),font=\itshape] \item  quantitative tracking accuracy in static and dynamic docking scenarios, \item  qualitative robustness compared to a standard onboard vision-based baseline, and \item  feasibility within the strict constraints of nano-scale aerial platforms\end{enumerate*}. 
All experiments compare the proposed method (Mag +\allowbreak Flow) against a baseline (Flow) that relies exclusively on the drone's native optical flow and internal state estimator.

\begin{table}[t]
\centering
\scriptsize
\setlength{\tabcolsep}{3pt}
\renewcommand{\arraystretch}{1.2}
\caption{\textbf{Scenario S1:} The UGV is static. For Hovering, the UAV performs the sequence in \Cref{fig:s1_takeoff}, \Cref{fig:s1_hover}, \Cref{fig:s1_landing}. For the test \emph{in-out}, the sequence includes also a movement (\qty{0.6}{\meter}) outside the landing pad. In summary, the sequence is: takeoff, fly \qty{0.6}{\meter} ahead, fly back, landing.  Values represent RMSE in centimeter, success rate (SC) in percentage.}
\label{tab:s1_results}
\begin{tabularx}{\linewidth}{c *{4}{>{\centering\arraybackslash}X}}
\toprule
\multirow{3}{*}{\textbf{Test}} & \multicolumn{2}{c}{\textbf{Hovering}} & \multicolumn{2}{c}{\textbf{in-out}} \\
\cmidrule(lr){2-3} \cmidrule(lr){4-5}
& \textbf{Mag+Flow} & \textbf{Flow} & \textbf{Mag+Flow} & \textbf{Flow} \\
\midrule
\textbf{1}  & 1.14 & 4.56 & 6.93 & FAIL \\
\textbf{2}  & 0.80 & 4.55 & 6.21 & FAIL \\
\textbf{3}  & 1.79 & 2.56 & FAIL & FAIL \\
\textbf{4}  & 1.76 & 5.36 & 6.94 & FAIL \\
\textbf{5}  & 4.01 & 5.46 & 6.07 & FAIL \\
\textbf{6}  & 7.11 & 3.24 & 15.7 & — \\
\textbf{7}  & 8.11 & 1.28 & 6.03 & — \\
\textbf{8}  & 2.84 & 2.58 & 6.01 & — \\
\textbf{9}  & 11.6 & 6.93 & 5.40 & — \\
\textbf{10} & 11.5 & 0.65 & 5.79 & — \\
\midrule
\textbf{Mean} & 5.01 & 3.72 & 7.23 & N/A \\
\midrule
\textbf{SC} & 100\% & 100\% & 90\% & 0\% \\
\bottomrule
\end{tabularx}
\end{table}

\subsection{Static Hovering and Landing (S1)}
In the first scenario (S1), the UGV acts as a stationary docking pad. The aerial agent performs a takeoff, stabilizes at a reference altitude of \qty{45}{\cm}, and executes an autonomous landing. Quantitative results, summarized in \Cref{tab:s1_results}, demonstrate that the magnetic aiding effectively eliminates the position drift inherent in dead-reckoning solutions. 

\begin{figure}[H]
    \centering
    \includegraphics[width=1\linewidth]{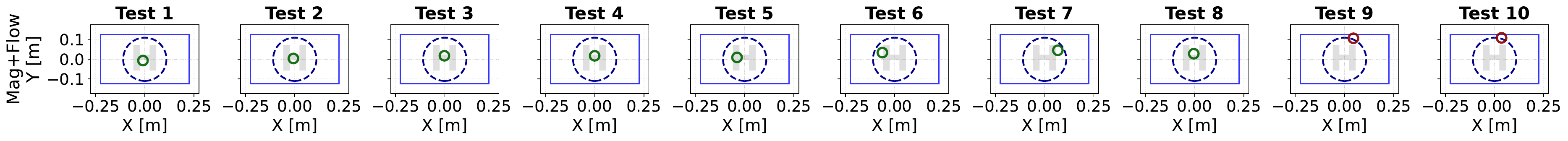}
    \caption{Touchdown accuracy analysis for scenario \textbf{S1}-Hovering. The plot shows the planar 2D position of the UAV relative to the docking pad center at the moment of landing for the proposed Mag+Flow method (circles).}
    \label{fig:Landing_qualitative-Results}
\end{figure}

The proposed system achieves a 3D RMSE below \qty{10}{\cm} for the Hovering sequence, with an aggregate mean of approximately \qty{5}{\cm}. This precision ensures that the UAV remains reliably centered over the \text{\diameter}\qty{22}{\cm} landing pad. Only in Test 9-10, the landing position is not perfectly centered on the landing pad, but not considered as \emph{failure}. 
\Cref{fig:Landing_qualitative-Results} visually illustrates the touchdown accuracy achieved by the proposed Mag+Flow method, with landing points tightly concentrated around the setpoint. For completeness,~\Cref{tab:s1_results} reports the corresponding quantitative results, including the Optical-Flow-only baseline for comparison. 
Any slight differences in favor of the Flow deck are likely within the variability of the touchdown transient (e.g., small oscillations during contact and ground effect induced disturbances) rather than reflecting a systematic advantage of the sensing modality.

This aspect is evident in a slightly different experiment, in fact, in addition to the nominal hover-and-land trials, we performed an extra out-and-back lateral offset maneuver (reported as \emph{in-out} in \Cref{tab:s1_results}): with the UGV still stationary, the UAV takes off, translates along the $x$ direction to a setpoint displaced by \qty{0.6}{\meter} from the docking-frame origin, and then returns toward the pad. This additional test stresses drift accumulation away from the docking region before the landing phase.
In these \emph{in-out} trials, the Flow-only baseline repeatedly violates the safety geofence ($>$\qty{0.5}{\meter} error) before a landing can be attempted, whereas Mag+Flow remains bounded and enables the subsequent precision landing keeping the 3D RMSE below \qty{10}{\cm} in most of the attempts.

\begin{figure}[H]
    \centering
    \includegraphics[width=1\linewidth]{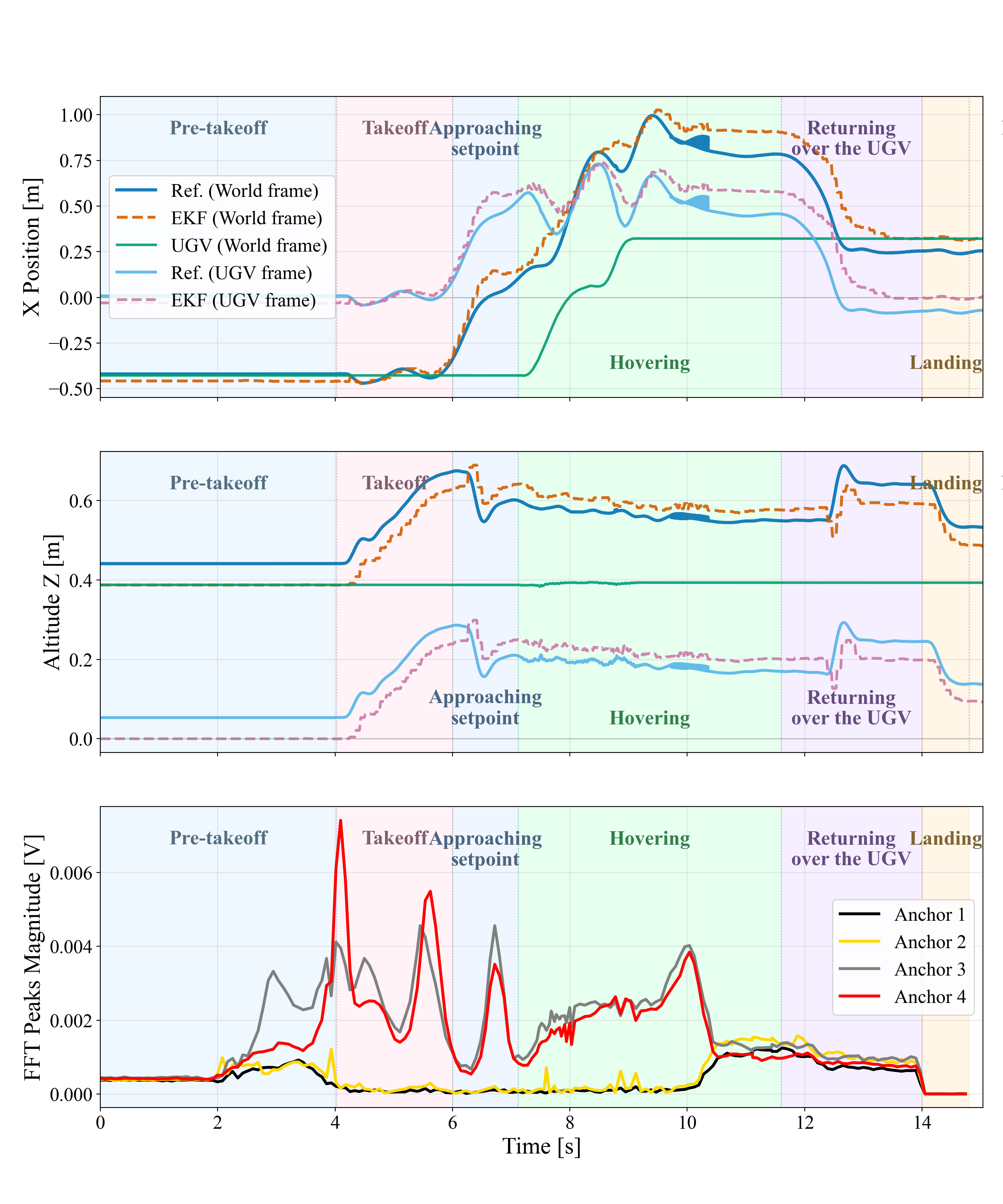}
    \caption{Scenario \textbf{S2}: time-history qualitative results for a representative linear docking sequence. The colored bands highlight the mission phases (pre-takeoff, takeoff/approach, hovering, return over the UGV, landing). Top: $x$ position in the world frame. Middle: altitude $z$. Bottom: magnitude of the FFT peaks of the coil voltage, which provides a proxy of the received magnetic signal strength over time.}
    \label{fig:S2}
\end{figure}

\subsection{Dynamic Tracking and Docking (S2 \& S3)}
The system's capabilities were further stressed in dynamic scenarios where the UGV executes linear back-and-forth motions (\textbf{S2}) and complex composite planar trajectories (\textbf{S3}).

\begin{table}[t]
\centering
\scriptsize
\setlength{\tabcolsep}{4pt}
\renewcommand{\arraystretch}{1.2}
\caption{Linear tracking and composite motion results. Performance comparison during UGV planar forward-backward motion (\textbf{S2}) and during complex UGV trajectories (\textbf{S3}). RMSE calculated on the whole path. Values represent RMSE in centimeter, success rate
(SC) in percentage.}
\label{tab:s2s3_results}
\begin{tabularx}{\linewidth}{c *{12}{Y}}
\toprule
 & \textbf{1} & \textbf{2} & \textbf{3} & \textbf{4} & \textbf{5} & \textbf{6} & \textbf{7} & \textbf{8} & \textbf{9} & \textbf{10} & \textbf{Mean} & \textbf{SC} \\
\midrule
\textbf{S2} & 6.04 & 14.7 & 7.11 & 6.09 & 10.7 & 5.56 & 13.5 & FAIL & 6.51 & FAIL & 8.77 & 80\% \\
\textbf{S3} & 9.39 & 8.14 & 27.1 & 7.14 & 11.3 & 7.09 & 7.34 & 8.52 & 9.53 & 9.46 & 10.5 & 100\% \\
\bottomrule
\end{tabularx}
\end{table}

\begin{figure}[H]
    \centering
    \includegraphics[width=\linewidth]{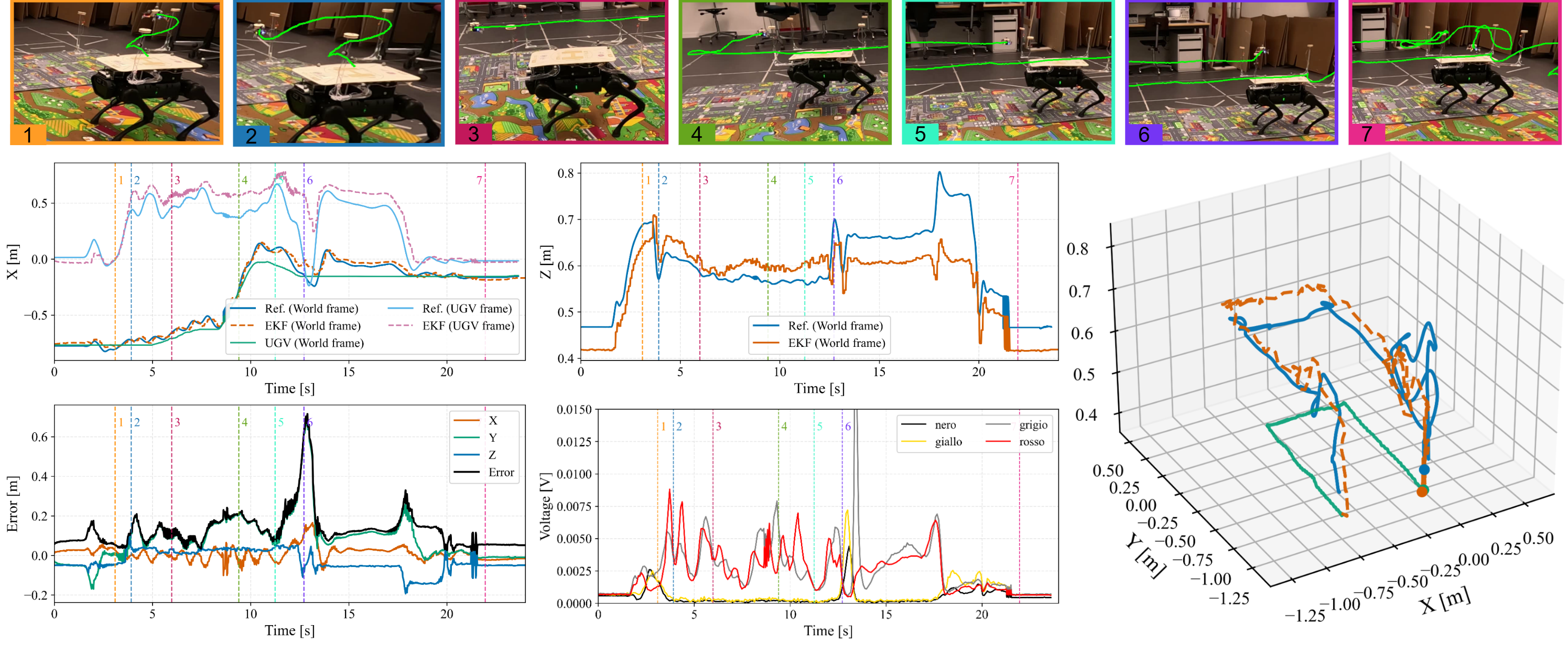}
    \caption{Scenario S3 (Test~1): qualitative time-aligned analysis of dynamic tracking. The numbered markers (1--7) identify key flight phases via synchronized video snapshots and the corresponding dashed vertical lines in the time traces. The trajectory plots (3D and top-down) compare \textit{Ref. (World frame)}, \textit{UGV (World frame)}, and \textit{EKF (World frame)}. Markers~1--2 (early takeoff/settling) highlight small transient oscillations. 
}

    \label{fig:S3}
\end{figure}

Quantitative results, summarized in \Cref{tab:s2s3_results}, confirm that the onboard estimator tracks the moving reference frame with an average 3D RMSE in the range of \qtyrange{8}{11}{\cm} and a success rate between 80\% and 100\%, defining system accuracy and robustness, respectively. A representative trial for Scenario \textbf{S3} is shown in \Cref{fig:S3}, where the estimated trajectory is compared against the UGV reference in the world frame. Therefore, \Cref{tab:s2s3_results} does not represent the landing accuracy, as reported in \Cref{tab:s1_results}, but instead the RMSE on the whole trajectory (including landing) referred to the expected trajectory. \Cref{fig:S2} and \Cref{fig:S3} depict an example of the movement sequence used to calculate the RMSE in \Cref{tab:s2s3_results}.

\Cref{fig:S2} provides a detailed breakdown of the system's behavior during the linear docking sequence (S2) using the magnetic localization. 
The shaded bands segment the mission into intuitive phases (pre-takeoff, approach, hovering, return, and landing) and allow directly relating the estimator outputs to the flight state. The first two plots report the UAV position along the $x$ axis and the altitude $z$ over time, highlighting the approach to the setpoint and the subsequent return over the UGV. 
The third plot shows the power spectrum magnitude  extracted from the receiving-coil voltage: as the relative geometry changes during the maneuver, the received signal strength varies accordingly, providing an intuitive proxy of link quality throughout the sequence.

From this visual analysis, two key characteristics emerge. First, the 'before takeoff' pose often exhibits a slight static offset below 5 cm on the xy plane, highlighting minor residual imperfections in the initial magnetic calibration, a behavior that does not significantly affects the system robustness and performance. Second, while the planar trajectories show high repeatability and consistent tracking, oscillations are observable during the transient phases of takeoff and landing (\Cref{fig:S2}-Approaching setpoint \& Hovering). These fluctuations are primarily attributed to the tuning of the flight controller specifically attributed to the nano-drone stability and dynamics rather than to instability in the magnetic localization estimate.

For Scenario \textbf{S3}, \Cref{fig:S3} complements the quantitative metrics with a time-aligned qualitative view: numbered video snapshots (markers 1--7) are synchronized with the corresponding dashed vertical markers in the logged signals, enabling a phase-by-phase interpretation of the maneuver. In the 3D plot, the \textit{Ref. (World frame)} trajectory is shown in blue, the \textit{UGV (World frame)} trajectory in green, and the \textit{EKF (World frame)} estimate as an orange dashed line. Similar to S2, small oscillations are visible during the initial takeoff/settling (markers 1--2), while the tracking error remains bounded for most of the trial. Noticeable error increases occur mainly at trajectory transitions and, more critically, around marker 6 (approximately at $t \approx 13$~s), where the UAV crosses the UGV footprint and the downward ToF ground reference changes abruptly from the deck to the floor. 
As discussed in Sec.~\ref{subsec:sensor_fusion} , we mitigate this effect with a dedicated step-detection and smoothing logic in the ToF driver; nevertheless, the residual discontinuity can still be perceived by the controller as an abrupt vertical variation, temporarily exciting oscillations and producing a corresponding increase in the tracking error (consistent with the snapshot associated to marker 6).

\subsection{System Integration}
{\color{orange}\color{black} 
The experimental campaign validates the MI-based final docking stage,
which is the focus of this paper. The proposed magnetic estimate is
injected into the Crazyflie state estimator through the same measurement
interface used by other absolute or relative positioning sources. This
makes hierarchical integration with mid range modalities technically
straightforward at the software level. However, a full UWB-to-MI handoff
experiment is outside the scope of the present validation. In this work,
we therefore report only the closed loop performance obtained with the
magnetic layer and the native onboard sensing stack.
}

\subsection{UGV Attitude Sensitivity Analysis}
\label{sec:results_limitations}
{\color{orange}\color{black}
The real world experiments reported above mainly involve planar translations of the UGV and bounded heading variations. However, a legged robot is not an ideally level platform: yaw rotations, body roll induced by gait, and pitch variations on uneven terrain change both the positions and the magnetic axes of the transmitting anchors. Since the current onboard estimator solves a position only inverse problem using the nominal anchor frame $B_0$, uncompensated UGV attitude changes introduce a model mismatch between the magnetic field generated by the true anchor configuration and the one assumed by the estimator.
\\
To quantify this effect we performed a model based sensitivity analysis in simulation. The nominal anchor geometry and dipole model used by the onboard estimator were kept unchanged, while the true UGV body frame $B_t$ was rotated with respect to the nominal estimator frame $B_0$. Synthetic magnetic measurements were generated from the rotated true anchor configuration and then processed by the same position only inverse localization and filtering pipeline used in the onboard implementation. The simulated receiver was placed at the representative hovering target used in the experiments, approximately $1$ m in front of the UGV and $0.45$ m above the nominal frame, and measurement noise was injected with relative standard deviation $\sigma_{\mathrm{rel}}=0.02$. Each condition was evaluated over 120 samples. This analysis is not intended to replace physical validation on uneven terrain, but to estimate the bias induced by uncompensated UGV attitude errors.
\\
\begin{figure*}[t]
    \centering

    \begin{subfigure}[t]{0.98\textwidth}
        \centering
        \includegraphics[width=\textwidth]{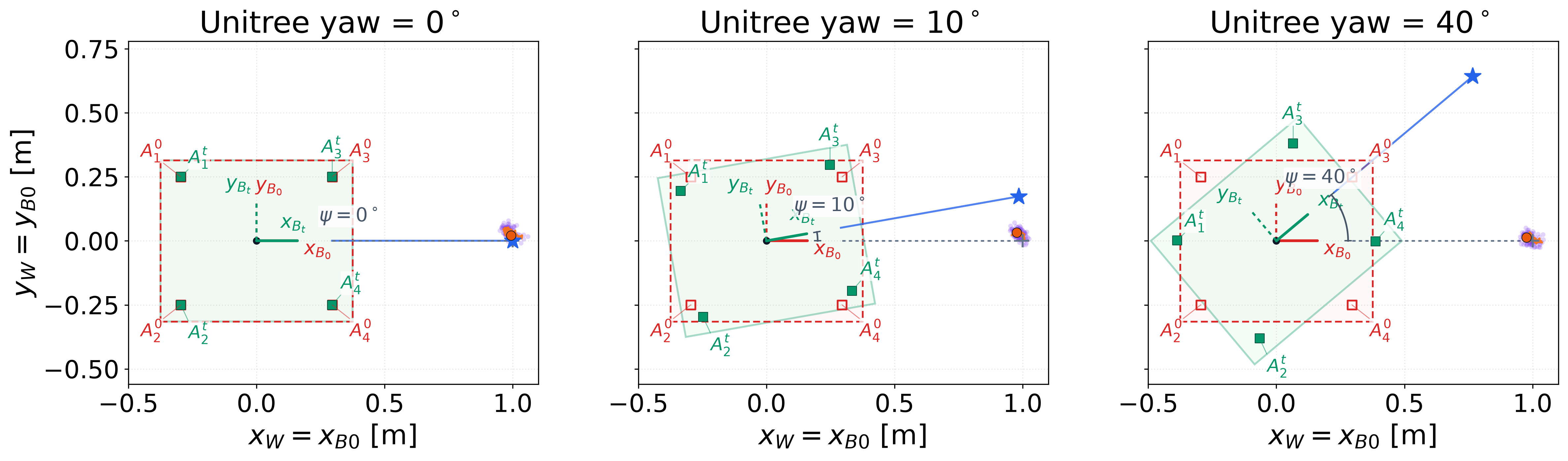}
        \caption{Yaw sensitivity. Uncompensated UGV yaw mainly induces horizontal bias in the estimated UAV position.}
        \label{fig:attitude_sensitivity_yaw}
    \end{subfigure}


    \begin{subfigure}[t]{0.98\textwidth}
        \centering
        \includegraphics[width=\textwidth]{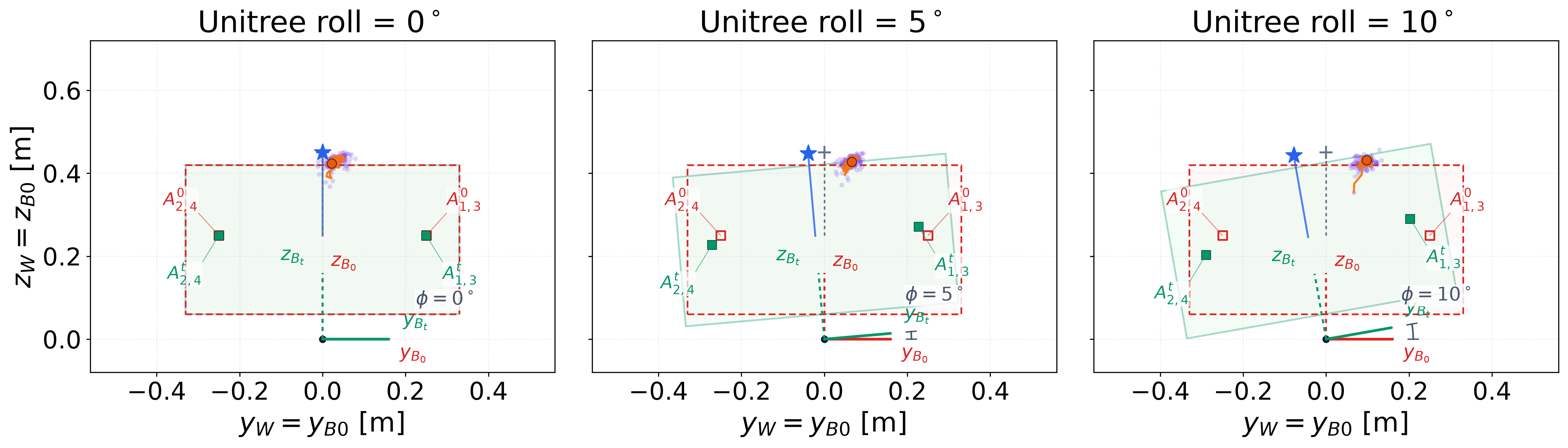}
        \caption{Roll sensitivity. UGV roll affects the lateral--vertical geometry between the anchors and the UAV.}
        \label{fig:attitude_sensitivity_roll}
    \end{subfigure}


    \begin{subfigure}[t]{0.98\textwidth}
        \centering
        \includegraphics[width=\textwidth]{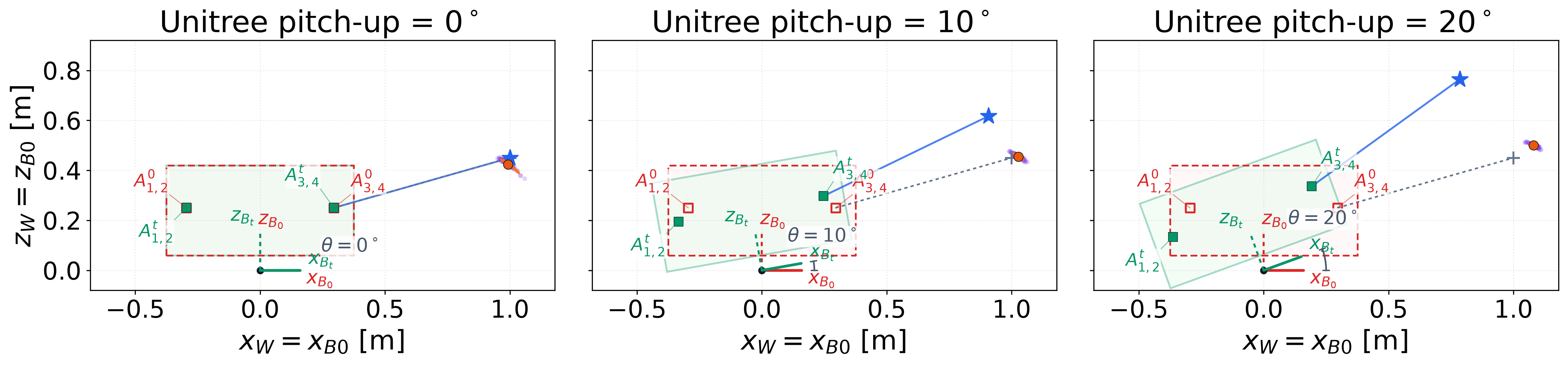}
        \caption{Pitch sensitivity. UGV pitch is particularly critical in the considered geometry because it changes both the forward and vertical relative anchor configuration.}
        \label{fig:attitude_sensitivity_pitch}
    \end{subfigure}


    \includegraphics[width=1\textwidth]{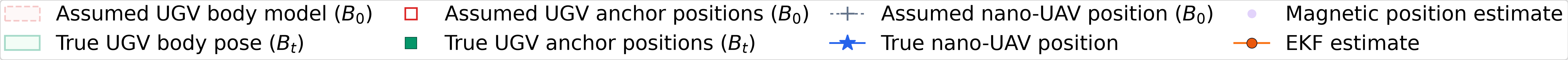}

    \caption{Model-based sensitivity of the magnetic localization pipeline to uncompensated UGV attitude. 
    The comparison shows that uncompensated UGV yaw, roll, and pitch are mapped by the current position-only estimator into systematic position biases.}
    \label{fig:ugv_attitude_sensitivity}
\end{figure*}

Fig.~\ref{fig:ugv_attitude_sensitivity} illustrates the resulting behavior for representative yaw, roll, and pitch rotations. The red dashed body and red square markers represent the nominal anchor geometry assumed by the estimator, while the green body and green markers represent the true rotated UGV body and anchor positions. The blue star indicates the true Crazyflie position in the world frame, the purple points show the magnetic position inputs, and the orange point is the final filtered estimate.
\\
The results show that uncompensated UGV attitude changes are interpreted primarily as position errors. In the nominal case, the final 3D error is approximately $3.45$ cm, consistent with the static noise level observed in the simulated pipeline. 
A yaw rotation of $10^\circ$ increases the final 3D error to $14.15$ cm, while a $40^\circ$ yaw produces a large horizontal bias of $66.38$ cm. Roll has a milder but still non negligible effect: $5^\circ$ roll increases the final 3D error to $10.81$ cm, and $10^\circ$ roll to $17.66$ cm. Pitch is the most critical in this configuration, because it directly changes the relative vertical geometry between the anchors and the receiver: $10^\circ$ pitch yields $20.59$ cm final 3D error, and $20^\circ$ pitch yields $39.91$ cm.
\\
\begin{table}[t]
\centering
\caption{model based sensitivity of the position only magnetic estimator to uncompensated UGV attitude. Errors are computed in the world frame. The projected error is reported in the dominant affected plane: $xy$ for yaw, $yz$ for roll, and $xz$ for pitch.}
\label{tab:ugv_attitude_sensitivity}
\begin{tabular}{lcccc}
\toprule
Motion & Angle & Final projected error & Final 3D error & Mean 3D error \\
 & [deg] & [cm] & [cm] & [cm] \\
\midrule
Yaw   & 0  & 2.28  & 3.45  & 4.51  \\
Yaw   & 10 & 14.05 & 14.15 & 14.76 \\
Yaw   & 40 & 66.33 & 66.38 & 67.38 \\
\midrule
Roll  & 0  & 3.39  & 3.45  & 4.51  \\
Roll  & 5  & 10.74 & 10.81 & 10.76 \\
Roll  & 10 & 17.59 & 17.66 & 17.30 \\
\midrule
Pitch & 0  & 2.68  & 3.45  & 4.51  \\
Pitch & 10 & 20.17 & 20.59 & 20.21 \\
Pitch & 20 & 39.64 & 39.91 & 40.03 \\
\bottomrule
\end{tabular}
\end{table}
\\
\\
These results refine the operational envelope of the current prototype. The experimentally validated system should be interpreted as an infrastructure-less close range localization solution for predominantly planar UGV motion and moderate body attitude variations. In contrast, operation on uneven terrain, aggressive turning in place, or large uncompensated body rotations can introduce errors exceeding the landing tolerance. This is not a severe practical limitation during the final touchdown phase, since a safe docking protocol should in any case command the UGV to stop, or at least minimize its body motion, before physical contact with the UAV. Therefore, the main impact of uncompensated UGV attitude is on the tracking and approach phases, rather than on the final landing procedure itself. Importantly, this limitation is not intrinsic to magnetic localization. 
If the UGV attitude is available from the quadruped IMU or communicated to the UAV, the same position only solver can be extended by rotating the anchor positions and dipole axes according to $R_{B_t}^{B_0}$. 
In that case, the inverse problem remains three dimensional in position, avoiding the computational cost of a full magnetic SE(3) estimator. 
A full 6-DoF magnetic pose solver would only be required when the UGV attitude is unknown or cannot be communicated reliably.
}
 
\section{Discussion}
\label{sec:discussion}

The experimental validation confirms that near field magneto inductive localization can serve as a robust, infrastructure-less navigation bridge for nano UAVs operating in close proximity to a mobile ground base. A key advantage highlighted by the comparison with the baseline (\Cref{sec:results}) is the immunity of the magnetic system to visual aliasing. In dynamic docking scenarios, standard infrastructure-less state estimators fails because the local frame aligns with the static global frame rather than the moving platform, leading to uncontrolled drift. Conversely, the proposed MI system locks the UAV's position relative to the field generated by the UGV anchors. This physical link ensures that the relative position estimate remains consistent regardless of the surrounding texture, lighting conditions, or the global motion of the ground robot. 
In contrast to permanent magnet docking approaches, where magnetics mainly provide passive short range capture, our frequency division MI beacons define a structured measurement model for relative navigation: anchor contributions are separable in the frequency domain, enabling unambiguous data association and online link quality monitoring while attenuating DC/low-frequency disturbances.
In line with this, \Cref{fig:S2} qualitatively shows how the received magnetic signal  evolves across the flight phases, providing a direct indication of link quality during the maneuver. This feature is critical for operations in unstructured environments (e.g., caves or disaster sites) where GNSS and external illumination are unavailable.

The system architecture was designed to handle the extreme asymmetry between the two robotic platforms. By shifting the power hungry components (transmitting coils and drivers) to the quadruped, which has orders of magnitude more battery capacity, and keeping the UAV side passive and lightweight, we respect the strict SWaP constraints of the nano UAV.
The receiving hardware adds less than \qty{9}{\gram} to the UAV's mass, and the full onboard workflow (\Cref{fig:overview_sistema_final_v5}) runs at \qty{20}{\hertz} in real time without impacting the other nano UAV MCU tasks. This efficiency implies that the solution is scalable to even smaller platforms or swarms, provided that the UGV can host the emitters. 
{\color{orange}\color{black}
The attitude sensitivity analysis in \Cref{sec:results_limitations} highlights an important distinction between UAV-side and UGV-side rotations. UAV roll and pitch are already compensated through the onboard attitude estimate used to rotate the receiving-coil normal. UGV attitude, instead, changes the magnetic map itself by rotating both the anchor positions and their dipole axes. The current implementation assumes this map to be fixed in the nominal docking frame, which is adequate for the experimentally validated planar tracking and landing maneuvers, but not for aggressive turning in place or operation on uneven terrain without attitude compensation. 

From a practical docking perspective, however, this is not a severe limitation during the final touchdown phase. On uneven terrain or during highly dynamic locomotion, a safe landing protocol should command the UGV to stop, or at least minimize body motion, before physical contact with the UAV. Therefore, uncompensated UGV attitude mainly limits the tracking and approach phases, while the final landing can still be executed under a quasi static UGV condition. Future implementations can further relax this constraint by communicating the quadruped IMU attitude to the UAV and rotating the anchor positions and dipole axes online, preserving the current position only magnetic solver without requiring a full magnetic SE(3) estimator.
}

\section{Conclusion and Future Work}
\label{sec:conclusion}

This paper presented the design and validation of a fully onboard, infrastructure-less MI localization system enabling heterogeneous interaction between a nano UAV and a quadrupedal robot. By leveraging frequency-multiplexed magnetic beacons, we demonstrated that a \qty{47}{\gram} drone can autonomously hover, track, and land on a moving platform with centimeter level precision, independent of GNSS or external motion capture systems.

The experimental results show that the system achieves a positioning RMSE of approximately \qty{5}{\cm} in static hovering and maintains an accuracy between \qtyrange{8}{11}{\cm} while tracking a moving UGV. In all field experiments, the success rate is always above 80\%. 
This performance significantly outperforms standard methods, listed in \Cref{tab:uav_ugv_landing_comparison}, which consistently fail in dynamic docking scenarios due to the lack of an absolute relative reference. Furthermore, the optimized implementation proves that complex nonlinear magnetic inversion is feasible in real time (\qty{20}{\hertz}) on resource-constrained robotic platforms.

This paper poses foundations for advancing the heterogeneous robotic cooperation in complex scenarios, proposing a solid system validated in the field with an UGV and a nano UAV.

\section*{Supplementary Material}
Supplementary video at \url{https://www.youtube.com/watch?v=rqQ7lGnRazo} \\ The hardware design files, firmware, and analysis scripts will be made
publicly available upon publication.

\section*{Declaration of generative AI and AI-assisted technologies in the manuscript preparation process}

During the preparation of this work, the authors used generative AI (OpenAI ChatGPT and Google Gemini) and AI-assisted technologies (GitHub Copilot and Codex) to support English-language revision, improve clarity and readability of the manuscript, and assist in the
writing of auxiliary Python code for data processing and plot generation.
After using these tools, the authors reviewed and edited the content as needed and take full responsibility for the content of the published article.

\section*{Funding}
This work has been supported by the European Commission HORIZON project RADIANCE, ID: 101235536, DOI: \url{https://doi.org/10.3030/101235536}.

\bibliographystyle{elsarticle-num}
\bibliography{references}

\end{document}